\documentclass[11pt,letterpaper]{article}

\usepackage[title]{appendix}

\usepackage{caption}
\usepackage{subcaption}
\usepackage{fullpage}
\usepackage{times}
\usepackage{array}
\usepackage{ragged2e}
\usepackage{multirow}

\usepackage[margin=1in]{geometry}

\usepackage{fancyhdr, amssymb, mathtools}
\usepackage[ruled,vlined]{algorithm2e}
\usepackage[dvipsnames,table]{xcolor}
\usepackage[margin = 1 in]{geometry}
\usepackage{graphicx}
\usepackage{subcaption}
\usepackage{outlines}
\usepackage{amsmath}

\usepackage{graphicx}
\usepackage{dirtytalk}

\usepackage{multirow}
\usepackage{arydshln}

\usepackage[english]{babel} 

\usepackage [autostyle, english = american]{csquotes}

\MakeOuterQuote{"}

\usepackage{amsmath}
\DeclareMathOperator{\Tr}{Tr}

\usepackage{blindtext}
\usepackage{amsfonts}
\usepackage{amssymb}
\usepackage{bbm}
\usepackage{mathtools}

\usepackage{hhline}
\usepackage{makecell}

\usepackage{xltabular}
\usepackage{arydshln} 
\usepackage{wrapfig}
\usepackage{enumitem}
\usepackage{floatrow}
\usepackage{xfrac}
\usepackage[font={footnotesize}, labelfont={bf}, labelsep=space, justification=justified, skip=0pt]{caption}
\usepackage[hang,flushmargin]{footmisc} 
\usepackage[colorlinks=true, allcolors=blue]{hyperref}

\usepackage{booktabs}

\definecolor{headcolor}{RGB}{255, 255, 255}
\definecolor{columncolor}{RGB}{255, 255, 255}
\definecolor{modelcolor}{RGB}{255, 255, 255}
\definecolor{optioncolor}{RGB}{255, 255, 255}

\definecolor{codegreen}{rgb}{0,0.6,0}
\definecolor{codegray}{rgb}{0.5,0.5,0.5}
\definecolor{codepurple}{rgb}{0.58,0,0.82}
\definecolor{backcolour}{rgb}{0.95,0.95,0.92}

\usepackage{listings}
\lstdefinestyle{mystyle}{
    backgroundcolor=\color{backcolour},   
    commentstyle=\color{codegreen},
    keywordstyle=\color{magenta},
    numberstyle=\tiny\color{codegray},
    stringstyle=\color{codepurple},
    basicstyle=\ttfamily\footnotesize,
    breakatwhitespace=false,         
    breaklines=true,                 
    captionpos=b,                    
    keepspaces=true,                 
    numbers=left,                    
    numbersep=5pt,                  
    showspaces=false,                
    showstringspaces=false,
    showtabs=false,                  
    tabsize=2
}
\usepackage{cleveref} 
\usepackage{cite} 

\usepackage{listings}
\usepackage{xcolor}

\definecolor{codegray}{gray}{0.95}

\lstdefinestyle{mystyle}{
    backgroundcolor=\color{codegray},   
    commentstyle=\color{gray},
    keywordstyle=\color{blue},
    numberstyle=\tiny\color{gray},
    stringstyle=\color{purple},
    basicstyle=\ttfamily\footnotesize,
    breaklines=true,
    captionpos=b,
    keepspaces=true,
    numbers=left,
    numbersep=5pt,
    showspaces=false,
    showstringspaces=false,
    showtabs=false,
    tabsize=2
}

\AtBeginEnvironment{appendices}{\crefalias{section}{appendix}}

\newcommand\parens[1]{\mathopen{}\left(#1\right)\mathclose{}}

\newcommand{\cmt}[1]{} 

\newcommand{\xb}{\boldsymbol{x}}

\newcommand{\yb}{\boldsymbol{y}}

\newcommand{\si}{\mathrm{i}}
\newcommand{\sj}{\mathrm{j}}

\newcommand{\ti}{\textit{i}}
\newcommand{\tj}{\textit{j}}

\usepackage{titling}
\thanksheadextra{}{}
\thanksheadextra{}{} 
\usepackage{lipsum} 

\title{Learning Mappings in Mesh-based Simulations}
\date{\vspace{-5ex}}
\usepackage{authblk}
\author[1]{Shirin Hosseinmardi}
\author[1,2]{Ramin Bostanabad \thanks{Corresponding Author: Raminb@uci.edu}}
\affil[1]{Department of Mechanical and Aerospace Engineering, University of California, Irvine}
\affil[2]{Department of Civil and Environmental Engineering, University of California, Irvine}

\begin{document}
\include{pythonlisting}
    \pagenumbering{arabic}
    \sloppy
    \maketitle
    \noindent \textbf{Abstract}\\

Many real-world physics and engineering problems arise in geometrically complex domains discretized by meshes for numerical simulations. The nodes of these potentially irregular meshes naturally form point clouds whose limited tractability poses significant challenges for learning mappings via machine learning models. To address this, we introduce a novel and parameter-free encoding scheme that aggregates footprints of points onto grid vertices and yields information-rich grid representations of the topology. Such structured representations are well-suited for standard convolution and FFT (Fast Fourier Transform) operations and enable efficient learning of mappings between encoded input-output pairs using Convolutional Neural Networks (CNNs). Specifically, we integrate our encoder with a uniquely designed UNet (E-UNet) and benchmark its performance against Fourier- and transformer-based models across diverse 2D and 3D problems where we analyze the performance in terms of predictive accuracy, data efficiency, and noise robustness. Furthermore, we highlight the versatility of our encoding scheme in various mapping tasks including recovering full point cloud responses from partial observations. Our proposed framework offers a practical alternative to both primitive and computationally intensive encoding schemes; supporting broad adoption in computational science applications involving mesh-based simulations.

    \section{Introduction} \label{sec intro}
\begin{figure}[!t]
    \centering
    \includegraphics[width=1\textwidth]{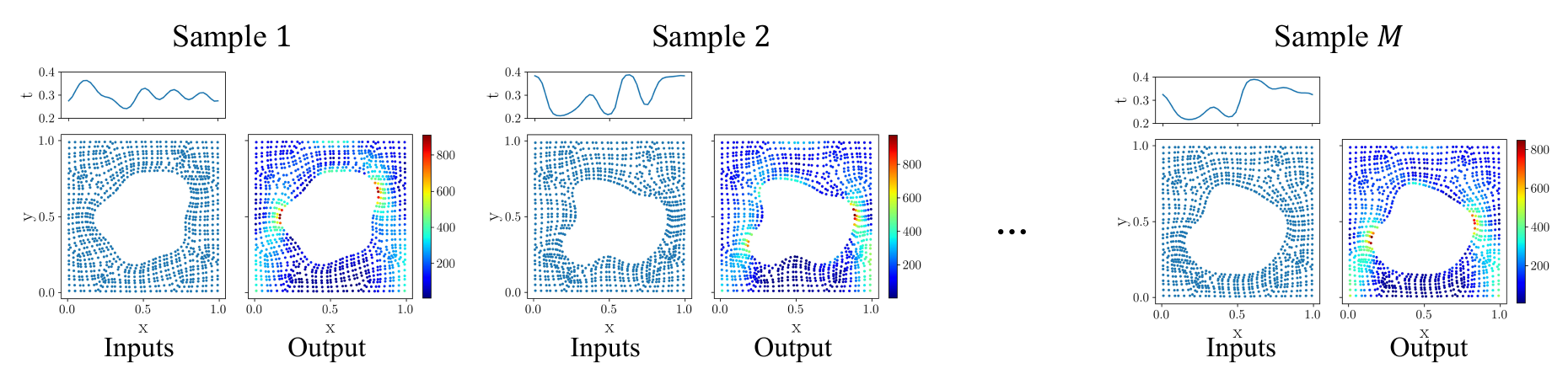} 
    \caption{\textbf{Training data from mesh-based simulations:} The response of a hyperelastic solid body to an external load is obtained via the finite element method (FEM). The solid body shape (and hence the mesh) as well as the applied load change across the samples. See \Cref{sec results} for more details on this benchmark problem.}
    \label{fig mesh data example}
\end{figure}

Mesh-based simulations are indispensable for modeling complex physical phenomena in various scientific and engineering disciplines. These simulations can generate large training datasets for building machine learning (ML) models that can surrogate physics-based simulations and, in turn, accelerate expensive tasks such as uncertainty quantification and inverse design. However, leveraging mesh-based data in ML is challenged by its intrinsic irregularity, i.e., the fact that observation and/or feature locations do not lie on a fixed regular grid (e.g., a Cartesian one), see \Cref{fig mesh data example} for an example. Our goal in this paper is to develop a computationally efficient method for learning input-output relations from mesh-based data or point clouds\footnote{Since our approach treats both mesh-based datasets and point clouds in the exact same way, we use these two terms interchangeably in this manuscript.}.

We broadly categorize the relevant works into two groups: 
(1) techniques that first map the data onto a regular grid where one can use conventional ML architectures (e.g., convolution- or Fourier-based); (2) methods that directly operate on the data, e.g., graph-, PointNet-, or transformer-based architectures. 
Most earlier works belong to the first category which enables the use of operations with highly optimized hardware support \cite{9536345}. For instance, early surrogate models for flow prediction rasterized geometry onto regular grids and employed convolutional neural networks (CNNs) \cite{guo2016convolutional,jacob2021deep}. Subsequent studies extended this idea to more complex flows and time-dependent problems, demonstrating that CNNs (especially UNet \cite{ronneberger2015u} variants) can successfully learn to predict physical fields from grid-based encodings of the domain \cite{obiols2021surfnet,wu2023computationally}. While sparse and continuous variants of convolutional operations have also been proposed \cite{liu2015sparse,xu2021ucnn,wang2018deep}, 
their adoption in scientific ML remains limited due to discretization errors and increased computational cost \cite{wang2018deep}. 

The encoding schemes adopted in the works that fall into the first category are quite basic and prone to information loss. Binary encoding that reduces a complex mesh to a boolean presence/absence indicator discards fine topological details \cite{le2022u,rezasefat2023finite}. Similarly, count-based and density rasterization (e.g., point count per grid cell) struggle with complex geometries or multi-scale phenomena unless an impractically fine grid is used \cite{devillers2000geometric,jo2020learning,esposito2024subdle,choy2025factorized,song2022dgpolarnet}. To address these limitations, more sophisticated encoding schemes with extensive preprocessing have been explored to better preserve information. For instance, Voronoi tessellation-based interpolation is implemented to map sparse fluid sensor data onto a grid for CNN reconstruction \cite{fukami2021global,cheng2024efficient}. Alternatively, space-filling curves can be employed to reorder mesh nodes and span the domain, suitable for 1D convolution \cite{heaney2024applying}. Coordinate transformation techniques are also utilized to map irregular domains onto structured reference grids or latent representations, although they typically rely on strict boundary alignment and require solving an auxiliary set of equations \cite{gao2021phygeonet,casenave2023mmgp}.

Both ends of the encoding spectrum, namely, naive binary/count-based encodings and overly parametrized pipelines, highlight the information-theoretic trade-off at play where advanced methods reduce information loss through intricate but often cumbersome and application-dependent implementations.
In this paper, we design a lightweight encoding scheme (detailed in \Cref{subsec encoding}) that is versatile and scalable while maximally preserving the information content.

In parallel with CNN-based approaches, the Fourier Neural Operator (FNO) has emerged as a benchmark spectral method for learning mappings between function spaces \cite{li2020fourier}. FNO lifts input functions into a latent space and applies learned global convolutions via FFT which inherently assumes a regular grid. Recent FNO variants generalize their applicability beyond regular geometries by transforming mesh node coordinates into pseudo-grids \cite{JMLR:v24:23-0064} or constructing a set
of domain-specific Fourier basis functions \cite{lingsch2023beyond}. Despite these advances, FNO variants suffer from spectral bias due to the practice of truncating high-frequency modes, which can filter out sharp or localized information \cite{khodakarami2025mitigating,pgcan2024}. For example, FNO variants fail to resolve the high-frequency variations near bubble interfaces \cite{hassan2023bubbleml}. As detailed in \Cref{sec results}, these results are consistent with our findings.

In the second category one of the most prominent line of research is perhaps based on graph neural networks (GNNs) which are naturally compatible with mesh-based data \cite{hamilton2017inductive,pfaff2020learning,deshpande2023convolution,seo2023graph,shivaditya2022graph}. However, GNNs impose topological assumptions about information flow, that is, interactions are local to the given adjacency. This assumption may not always hold as further modifications to the graph structure might be required to promote long-range dependencies in some cases \cite{krokos2024graph}. There are also cases where the connectivity information is redundant altogether, and the mesh nodes alone may suffice to capture the underlying topology \cite{casenave2023mmgp}.
Moreover, inference on GNNs often entails iterative message passing
across many edges, which is less memory- and compute-efficient on GPUs compared to the batched convolution operations \cite{zhang2022understanding}. Therefore, we forsake explicit graph representations and treat the simulation mesh nodes as potentially unstructured point clouds.

Treating mesh nodes as point clouds has long been investigated in computer vision tasks such as image segmentation and classification \cite{qi2017pointnet,qi2017pointnet++}, and has recently gained popularity in computational mechanics. For instance, PointNet variants are leveraged to predict laminar flow field around objects of different geometries \cite{kashefi2021point,kashefi2025kolmogorov}. PointNets are also trained using physics-informed loss functions to predict the displacement field in a linear elastic material with holes of different shapes \cite{kashefi2023physics}.
Despite their edge-free nature and avoidance of interpolation, these models lack an explicit mechanism to incorporate sources of variation beyond domain topology, such as boundary or initial conditions and variable input fields. Furthermore, PointNet-based models tend to converge slower than convolution- and graph-based counterparts, due in part to weaker inductive biases and the absence of spatial locality \cite{liu2022tree}.

Transformer-based models, initially developed for natural language processing (NLP), have also shown great promise in handling unstructured data for scientific computing tasks \cite{hao2023gnot,zhao2023pinnsformer,li2022transformer}. Unlike GNNs where positions are only relatively defined in the context of $k$-hop neighborhood, absolute positions of nodes are often incorporated in transformer-based models, making them explicitly position-aware \cite{cheng2025machine}. The self-attention mechanism, as a core component of such models, allows for selectively focusing on different parts of an input sequence and learning both short-range and long-range dependencies. However, attention complexity scales quadratically with the sequence length, which is a critical bottleneck when applied to simulation meshes comprising thousands of nodes \cite{NIPS2017_3f5ee243}. As demonstrated in our experiments with large point clouds in \Cref{sec results}, this computational overhead results in significantly longer training times compared to CNN- and Fourier-based models.

Our proposed method belongs to the first category where we develop an optimal encoding method that maps mesh-based data and point clouds to a grid representation that facilitates learning via a uniquely designed UNet architecture. More specifically, our contributions are as follows: 
(1) We propose an information-rich, fast, and parameter-free encoding scheme to represent point cloud data (with arbitrary dimensions) on structured grids with desirable resolutions.
(2) We develop a customized UNet model (E-UNet) to learn from mesh-based data. 
(3) We evaluate the performance of our approach against various competing methods, including Fourier- and transformer-based baselines, to showcase its mapping approximation accuracy, data efficiency, and noise robustness. 
(4) We provide a rigorous error and complexity analysis that describes the error and scaling behavior of our framework. 
(5) We demonstrate that our encoded topology and response representations can be used in a variety of mapping tasks, e.g., missing points' response recovery. Our
codes and datasets are publicly available on GitHub.

The remainder of this paper is organized as follows. 
We propose our methodology in \Cref{sec method} where we introduce our encoding scheme, analyze its encoding capacity from an information theoretic perspective, and also develop a fast reconstruction procedure. We describe the customized UNet architecture and its training strategy in \Cref{subsec unet}. In \Cref{subsec error,subsec complexity}, we provide theoretical estimates for error and time complexity of our framework (E-UNet). \Cref{sec results} presents a comprehensive evaluation of E-UNet through quantitative and qualitative comparisons against baselines across canonical 2D and 3D problems, followed by sensitivity studies that analyze the effects of noise and training data size. 
We provide concluding remarks and future research directions in \Cref{sec conclusion}.
    \section{Proposed Methodology} \label{sec method}

Effective representation and processing of mesh-based point clouds is pivotal to our objective of predicting the responses at their scattered coordinates given the raw point cloud and optionally additional inputs such as initial or boundary conditions (ICs or BCs). This task requires an encoding scheme that preserves fine topological details while admitting highly optimized grid-based operations, such as convolutions which serve as powerful hierarchical feature extractors. To achieve this, we develop a framework that has three primary modules: 
(1) a compact and parameter-free encoder that rasterizes the point cloud data onto a regular grid with a desirable resolution that balances reconstruction error and computational costs of training an ML model, 
(2) a reliable reconstruction mechanism to approximate the point cloud responses with negligible yet quantifiable error, and
(3) a unique UNet architecture that is trained to learn the grid-to-grid mapping between the encoded topology and other inputs (e.g., ICs and/or BCs) with the encoded responses.

Below, we first introduce our approach for encoding the topology and responses where we also elaborate on the optimality of our encoding scheme and how it can be inverted, i.e., to do reconstruction. Then, we describe the UNet architecture that we have designed to exploit the representations that our encoding scheme provides. We conclude this section by providing theoretical error and computational complexity analyses.

\subsection{Analytic Encoding of Topology and Response Field} \label{subsec encoding}%

We propose an analytic grid-based encoding to rapidly transform point cloud data into structured representations suitable for our learning task. Without loss of generality, we assume each batch of data consists of $M$ point clouds. For notational simplicity, we further presume that each point cloud contains $N$ scattered points in the 2D plane (i.e., the number of points does not change across the samples). Extensions of our approach to 3D (with or without time) and to cases where the size of the point cloud changes across the samples is straightforward. These extensions are tested in \Cref{sec results}.

\subsubsection{Topology Encoding} \label{subsubsec encode}

\begin{figure}[!t]
    \centering
    \includegraphics[width=0.99\textwidth]{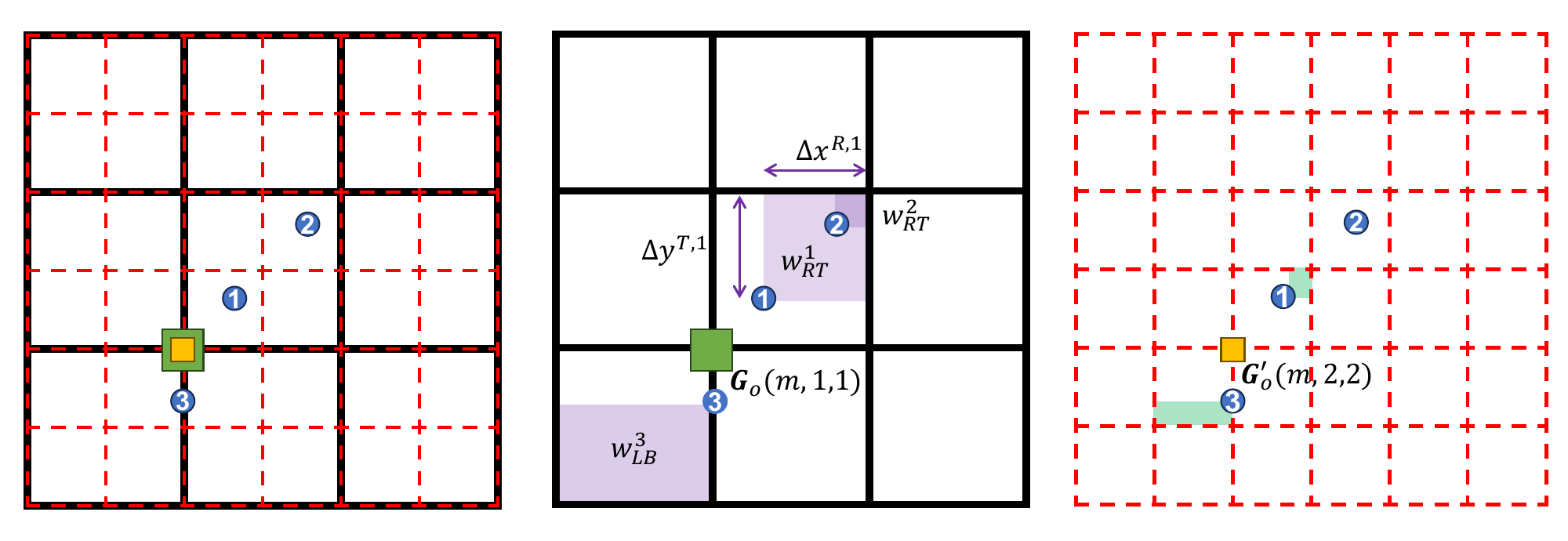} 
    \caption{\textbf{Topology encoding:} The domain is encoded with two resolutions: $r = 4$ for the coarse black grid and $r' = 7$ for the fine red grid. 
    $\boldsymbol{G}_o(m, 1,1)$ denotes the encoded value at vertex $(1, 1)$ for sample $m$ and is obtained by accumulating three weight values, i.e., $G_o(m, 1,1) = w^1_{RT} + w^2_{RT} + w^3_{LB}$. These weights are calculated based on the distance of vertex $(1, 1)$ from the three points ($N=3$ in this figure). 
    Shorter distances result into larger weights which are equal to the area of the shaded squares.}
    \label{fig topology encoding}
\end{figure}

Given a batch of $M$ point clouds with scaled coordinates $\boldsymbol{x} \in [-1,1]^{M  \times N}$ and $\boldsymbol{y} \in [-1,1]^{M  \times N}$, our topology encoder aims to produce the structured grid representation $\boldsymbol{G}_o \in \mathbb{R}^{M \times r \times r}$ where $r$ is the grid resolution (so $\boldsymbol{G}_o$ has $(r-1)\times(r-1)$ cells). As detailed below and schematically illustrated in \Cref{fig topology encoding}, our approach essentially assigns weights to vertices of a grid based on their distance from point $n$ in the point cloud. By repeating this process for $n=1, ..., N$ and summing the weight contributions to each vertex, we obtain $\boldsymbol{G}_o$. For simplicity, we have assumed that $\boldsymbol{G}_o$ has $r-1$ cells in each dimension but the following formulas can be easily extended to other settings. 


To calculate the vertex weights of each point in the point cloud, we need to identify the vertices of the cell that encloses each point. To this end, we first convert $\xb$ and $\yb$ to grid indices:
\begin{equation}
\begin{aligned}
\boldsymbol{i}_x = \cfrac{\boldsymbol{x} + 1}{2} (r - 1) \in \left[0, r-1\right]^{M  \times N}\\
\boldsymbol{i}_y = \cfrac{\boldsymbol{y} + 1}{2} (r - 1) \in \left[0, r-1\right]^{M \times N},
\end{aligned}
\label{eq grid indices continuous}
\end{equation}
and then obtain the indices of the four vertices of the enclosing cell via:
\begin{equation}
\boldsymbol{i}_x^L = \lfloor \boldsymbol{i}_x \rfloor, \quad \boldsymbol{i}_x^R =\boldsymbol{i}_x^L + 1, \quad \boldsymbol{i}_y^B = \lfloor \boldsymbol{i}_y \rfloor, \quad \boldsymbol{i}_y^T = \boldsymbol{i}_y^B + 1 \in \{0 ,..., r-1\}^{M \times N} , 
\label{eq grid indices discrete}
\end{equation}
where $\lfloor \cdot \rfloor$ is the floor function and superscripts $L,R,B,$ and $T$ denote left, right, bottom, and top, respectively. 
Each tensor in \Cref{eq grid indices discrete} contains the vertex coordinate of the enclosing cell of a point, e.g., $\boldsymbol{i}_x^L(m,n)$ for $n\in\{1, \cdots, N\}$ and $m\in \{1, \cdots, M\}$ denotes the grid index of the left edge of the cell that contains point $n$ in the $m^{th}$ point cloud. 
We highlight that the operations in \Cref{eq grid indices continuous,eq grid indices discrete} have minimal computational overhead which is one of the primary reasons we have chosen a regular grid for our encoding. 

Once the indices in \Cref{eq grid indices continuous,eq grid indices discrete} are identified, the weights are obtained by assuming that within each cell, the underlying topology field varies locally affinely in the $x$- and 
$y$ -directions.
We recognize that this assumption may introduce some error which we study further in \Cref{subsec error}. Under this assumption, the weights correspond to partitions of the unit cell, determined by the points’ relative distances to the edges of their enclosing cells. That is:

\begin{equation}\label{eq weights} 
\begin{aligned}
\boldsymbol{w}_{RB} =\boldsymbol{\Delta x}^R \boldsymbol{\Delta y}^B, \quad \boldsymbol{w}_{LB} =\boldsymbol{\Delta x}^L \boldsymbol{\Delta y}^B \in [0,1]^{M \times N}
\\
\boldsymbol{w}_{RT} =\boldsymbol{\Delta x}^R \boldsymbol{\Delta y}^T, \quad \boldsymbol{w}_{LT} =\boldsymbol{\Delta x}^L \boldsymbol{\Delta y}^T \in [0,1]^{M \times N}
\end{aligned}
\end{equation}
where:
\begin{equation}
\begin{aligned}
\boldsymbol{\Delta x}^R = \boldsymbol{i}_x^R - \boldsymbol{i}_x, \quad \boldsymbol{\Delta x}^L = 1 - \boldsymbol{\Delta x}^R,
\\
\boldsymbol{\Delta y}^T = \boldsymbol{i}_y^T - \boldsymbol{i}_y, \quad \boldsymbol{\Delta y}^B = 1 - \boldsymbol{\Delta y}^T,
\end{aligned}
\end{equation}
and $\boldsymbol{w}_{LT}(m, n)$ denotes the top left cell partition associated with point $n$ in the $m^{th}$ point cloud.

We define four sets of weights for each $\boldsymbol{G}_o(m , \ti , \tj)$ where $(\ti , \tj) \in \{0,...,r-1\} \times\{0,...,r-1\}$. Each set contains the relevant weights of the points in the adjacent cells that contribute to the  $(\ti , \tj)$ vertex. That is:
\begin{equation}
    \begin{aligned}
    \mathbb{W}_{LT}(m , \ti , \tj) = \{\boldsymbol{w}_{RB}(m,n) ~|~ \boldsymbol{i}_x(m,n) \in [\ti-1 , \ti]  \wedge \boldsymbol{i}_y(m,n) \in [\tj , \tj+1] \} \label{eq w_lt}
    \\
    \mathbb{W}_{LB}(m , \ti , \tj) = \{\boldsymbol{w}_{RT}(m,n)~|~  \boldsymbol{i}_x(m,n) \in [\ti-1 , \ti]  \wedge \boldsymbol{i}_y(m,n) \in [\tj-1 , \tj] \}
    \\
    \mathbb{W}_{RT}(m , \ti , \tj) = \{\boldsymbol{w}_{LB}(m,n) ~|~   \boldsymbol{i}_x(m,n) \in [\ti , \ti+1]  \wedge \boldsymbol{i}_y(m,n) \in [\tj , \tj+1] \}
    \\
    \mathbb{W}_{RB}(m , \ti , \tj) = \{\boldsymbol{w}_{LT}(m,n)~|~  \boldsymbol{i}_x(m,n) \in [\ti , \ti+1]  \wedge \boldsymbol{i}_y(m,n) \in [\tj-1 , \tj] \} ,
    \end{aligned}
\end{equation}
For example, $\mathbb{W}_{LT}(m , \ti , \tj)$ collects the weights $\boldsymbol{w}_{LT}$  of the points located in the top left cell of vertex $(\ti , \tj)$ corresponding to point cloud $m$.
Then, with $\mathbb W(m,\ti , \tj) = \bigcup_{a \in \{  LT , LB , RT , RB\}}  \mathbb W_a(m,\ti , \tj)$, we have:
\begin{equation}
\begin{aligned}
\boldsymbol{G}_o(m , \ti , \tj) = \sum_{w \in \mathbb{W}(m,\ti,\tj) }\ w,
\end{aligned}
\end{equation}
which represents accumulating weight contributions from surrounding points onto the vertex $(\ti,\tj)$ for point cloud $m$. Note that by convention, if $\mathbb W(m,\ti , \tj)$ is an empty set, $\boldsymbol{G}_o(m , \ti , \tj)$ is $0$. 

An attractive feature of our encoding scheme is that it preserves the total point count in the point cloud. This is because the contributions associated with each point $n\in\{1, \cdots, N\}$ in the point cloud sum to one (i.e., the weights are partitions of a unit cell) and consequently the summation of values at the grid vertices remains equal to $N$ (i.e., $\sum_{\ti = 0}^{r-1}\sum_{\tj = 0}^{r-1} \boldsymbol{G}_o(m , \ti , \tj) = N $).

\subsubsection{Optimality of the Proposed Topology Encoding} \label{subsubsec info}
A richer encoded representation captures more of the subtle topological details, enabling the ML model to more accurately learn to map the mesh (or point cloud) and input function(s) to the response field.
To demonstrate the richness (i.e., capacity) of our topology encoding, we compare its information content against two alternative schemes that are commonly used in the literature: binary encoding  \cite{le2022u,rezasefat2023finite} and count-based encoding \cite{jo2020learning,esposito2024subdle,choy2025factorized,song2022dgpolarnet}. In binary encoding, the vertices of a cell are assigned the value $1$ if the cell contains at least one point and $0$ otherwise.  Count-based encoding records the exact number of points per cell, preserving some information on points' density, rather than capturing only the cell occupancy. While learning-based encoding schemes also exist\cite{hamilton2017inductive,zhou2024text2pde}, their additional trainable parameters increase complexity and computational overhead, undermining our goal of maintaining a computationally efficient, versatile, and interpretable encoding.

The information content is defined as the minimum number of bits required to uniquely encode the state of the system. Formally, if the system is in any of the $|\Omega|$ equally likely configurations, then the information content is $\mathcal{H}=\log_2 |\Omega|$. Assuming the positional quantization is $\delta$, the information content of the original mesh or point cloud (i.e., maximum available information) is given by:
\begin{equation} \label{eq H_x}
    \mathcal{H}_{\boldsymbol{X}}= \log_2 \binom{4/ \delta^2}{N},
\end{equation}
which represents ways to place $N$ indistinguishable points into $\frac{4}{\delta^2}$ distinguishable positions in the $[-1 , 1]^2$ domain.

In binary encoding, for $q \in \{1, \dots,\min\parens{ N , (r-1)^2}\}$ occupied cells, there are $\binom{(r-1)^2}{q}$ occupancy patterns. Therefore, the information content can be written as:
\begin{equation} \label{eq H_b}
    \mathcal{H}_B= \log_2 \parens{\sum^{\min( N , (r-1)^2)}_{q=1} \binom{(r-1)^2}{q}}.
\end{equation}
Count-based encoding is a classic "stars and bars" problem, in that we look for the total number of distinct ways to distribute $N$ indistinguishable points over $(r-1)^2$ distinguishable cells. Thus, its information content is the following: 
\begin{equation} \label{eq H_c}
    \mathcal{H}_C= \log_2 \binom{N+(r-1)^2-1}{N} = \log_2 \binom{N+r^2 -2r}{N}.
\end{equation}
To derive the information content of our encoding, we recognize that the spatial precision of our topology encoding is strictly below machine precision (i.e., $\delta_{eff}^2 > \delta^2$) and we presume that it inherits the same error characteristics as bilinear interpolation. Hence, we model the effective spatial quantization as:
\begin{equation} \label{eq delta eff}
    \delta_{eff}^2 \approx \delta^2 + O(h^\beta),
\end{equation}
where $h = \frac{2}{r-1}$ is the grid spacing and $\beta \in [0,2]$ depends on the smoothness of the underlying topology field (i.e., the function that describes the spatial layout of the domain). We study the term $O(h^\beta)$ in detail in \Cref{subsec error}.  Assuming this function is twice continuously differentiable and has bounded second derivatives, we can roughly approximate the effective spatial quantization as  $\delta_{eff}^2 \approx \delta^2 + \mathcal{\eta}h^2$ for some constant $\eta$. The information content of our encoding is thus given by:
\begin{equation} \label{eq H_g}
    \mathcal{H}_{\boldsymbol{G}} = \log_2 \binom{4/ \delta_{eff}^2}{N} \approx \log_2 \binom{\frac{4(r-1)^2}{\delta^2(r-1)^2 + 4\eta}}{N}.
\end{equation}

A reasonable estimate for positional quantization is $\delta = 10^{-3}$ for $10$-bit fraction numbers. If the second derivatives of the topology field are $O(0.1)$, we have $\eta \approx 0.01$. 
We can evaluate the corresponding information contents by plugging $N = 2000,~ r = 128$ into \Cref{eq H_x,eq H_b,eq H_c,eq H_g}. Hence, $\mathcal{H}_{\boldsymbol X} , \mathcal{H}_B, \mathcal{H}_{C}$ and $\mathcal{H}_{G}$ are approximately $24.81, 8.72, 9.07$ and $21.21~kbits$, respectively. In this instance, $85\%$ of the available information is encoded by our topology encoding, whereas binary and count-based encodings retain only about $35\%$.
With $N=1000$ and all other parameters fixed, these percentages change to $87\%$ for our encoding and to $40\%$ and $41\%$ for binary and count-based encodings, respectively. This indicates that all methods capture more of the available information when tasked to encode fewer points. Nonetheless, our encoding scheme maintains a consistently higher information content.

\subsubsection{Response Encoding}

We encode the responses $\boldsymbol{U} \in \mathbb{R}_{+}^{M \times N}$ at point cloud coordinates $\boldsymbol{X} = [\boldsymbol{x} , \boldsymbol{y}] \in [-1,1]^{M \times N \times 2}$ via grid $\boldsymbol{G}_u \in \mathbb{R}^{M \times r \times r}$. Each point $n \in \{1,\cdots,N\}$ in point cloud $m \in \{1,\cdots,M\}$ distributes its response value $\boldsymbol{U}(m , n)$ to the four vertices of its enclosing cell based on the weights obtained via \Cref{eq weights}. 
Due to the cumulative nature of our approach, all response values must be non-negative. Thus, if the reference responses are negative, we first map them to $\mathbb{R}_+$ by either simply adding a constant to $\boldsymbol{U}$ or min-max normalizing the reference values $\boldsymbol{U}$.

Similar to topology encoding, we define four sets that represent the response contributions from points located in the cells adjacent to the vertex $(\ti,\tj)$ :
\begin{equation} 
\begin{aligned}
\mathbb{U}_{LT}(m , \ti , \tj) = \{\boldsymbol{w}_{RB}(m,n)\boldsymbol{U}(m,n) ~|~  \boldsymbol{i}_x(m,n) \in [\ti-1 , \ti]  \wedge \boldsymbol{i}_y(m,n) \in [\tj , \tj+1] \}
\\
\mathbb{U}_{LB}(m , \ti , \tj) = \{\boldsymbol{w}_{RT}(m,n)\boldsymbol{U}(m,n)~|~ \boldsymbol{i}_x(m,n) \in [\ti-1 , \ti]  \wedge \boldsymbol{i}_y(m,n) \in [\tj-1 , \tj] \}
\\
\mathbb{U}_{RT}(m , \ti , \tj) = \{\boldsymbol{w}_{LB}(m,n)\boldsymbol{U}(m,n) ~|~  \boldsymbol{i}_x(m,n) \in [\ti , \ti+1]  \wedge \boldsymbol{i}_y(m,n) \in [\tj , \tj+1] \}
\\
\mathbb{U}_{RB}(m , \ti , \tj) = \{\boldsymbol{w}_{LT}(m,n)\boldsymbol{U}(m,n)~|~ \boldsymbol{i}_x(m,n) \in [\ti , \ti+1]  \wedge \boldsymbol{i}_y(m,n) \in [\tj-1 , \tj] \}. \\
\end{aligned}\label{eq response encoding}
\end{equation}
Then, with $\mathbb U(m,\ti , \tj) = \bigcup_{a \in \{  LT , LB , RT , RB\}}  \mathbb {U}_a(m,\ti , \tj)$, we have:
\begin{equation}
\begin{aligned}
\boldsymbol{G}_u(m , \ti , \tj) = \sum_{\mu \in \mathbb{U}(m,\ti,\tj) }\ \mu.
\end{aligned}
\end{equation}
$\boldsymbol{G}_u$ is the weighted grid representation of the response field where each vertex $(\ti , \tj)$ accumulates weighted contributions from nearby point responses based on their relative positions within their enclosing cell.

\subsubsection{Response Field Reconstruction} \label{subsubsec rec}
The ML model in \Cref{subsec unet} operates in the encoded spaces of topology and response field so we develop a reconstruction algorithm to map the encoded response (or its predicted encoding) to the original space. 
More specifically, given the encoded topology $\boldsymbol{G}_o$ and the encoded responses $\boldsymbol{G}_u$ (or its approximation $\widetilde{\boldsymbol{G}}_u$ by the ML model), we aim to reconstruct the response values $\widehat{\boldsymbol{U}}$ (or $\widetilde{\boldsymbol{U}}$ if based on $\widetilde{\boldsymbol{G}}_u$) at the point cloud coordinates $\boldsymbol{X}$.
To this end, we normalize $\boldsymbol{G}_u$ by $\boldsymbol{G}_o$ to obtain $\boldsymbol{H} = \boldsymbol{G_u}/(\boldsymbol{G_o} +10^{-6})$ where the division is done element-wise and $10^{-6}$ is added to ensure numerical stability. 
This normalization ensures that the accumulated response values are converted into a proper weighted average. Then, the four nearest vertices' values of $\boldsymbol{H}$ are interpolated\footnote{Bilinear interpolation in 2D.} based on the pre-computed interpolation weights and indices:  
\begin{equation} \label{eq rec_u}
    \boldsymbol{\widehat U}(\boldsymbol{i}_{x} , \boldsymbol{i}_{y}) = \boldsymbol{w}_{LT} \odot \boldsymbol{H}(\boldsymbol{i}_{x}^{L} , \boldsymbol{i}_{y}^T) + \boldsymbol{w}_{RT} \odot \boldsymbol{H}(\boldsymbol{i}_{x}^{R} , \boldsymbol{i}_{y}^T)+
     \boldsymbol{w}_{LB} \odot \boldsymbol{H}(\boldsymbol{i}_{x}^{L} , \boldsymbol{i}_{y}^B) + \boldsymbol{w}_{RB} \odot\boldsymbol{H}(\boldsymbol{i}_{x}^{R} , \boldsymbol{i}_{y}^B)
    \in \mathbb{R}^{M\times N},
\end{equation}
where $\odot$ is the Hadamard product and $\boldsymbol{H}(\boldsymbol{i}_{x}^{L} , \boldsymbol{i}_{y}^T)$ corresponds to the values at the top-left vertices of the cells containing $N$ points, indexed by $(\boldsymbol{i}_{x}^{L}, \boldsymbol{i}_{y}^T)$ accounting for all $r\times r$ grids in $\boldsymbol{H}$.
This reconstruction introduces some error depending on the resolution and the smoothness of the underlying function. A detailed analysis of this error is given in \Cref{subsec error}.

Note that even though the above study assumes a constant number of points per point cloud, the same approach could be taken for point clouds of different sizes where each point cloud is processed separately to obtain its corresponding grid representations.

\subsection{Mapping Encodings via UNet} \label{subsec unet}
\begin{figure}[!b]
    \centering
    
    \begin{subfigure}{1.0\textwidth}
        \centering
        \includegraphics[width=\textwidth]{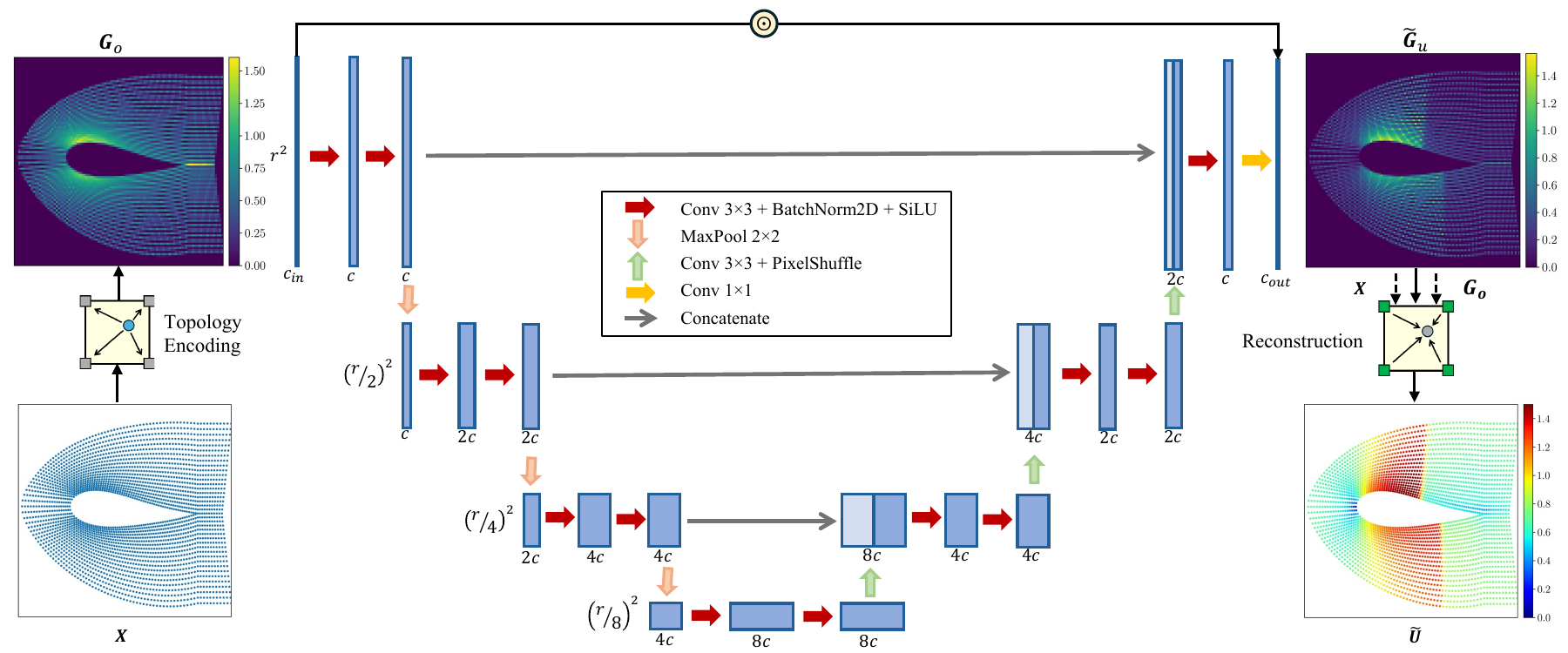}
        \caption{}
        \label{fig flowchart-a}
    \end{subfigure}
    
    \vspace{1em} 

    \begin{subfigure}{1.0\textwidth}
        \centering
        \includegraphics[width=\textwidth]{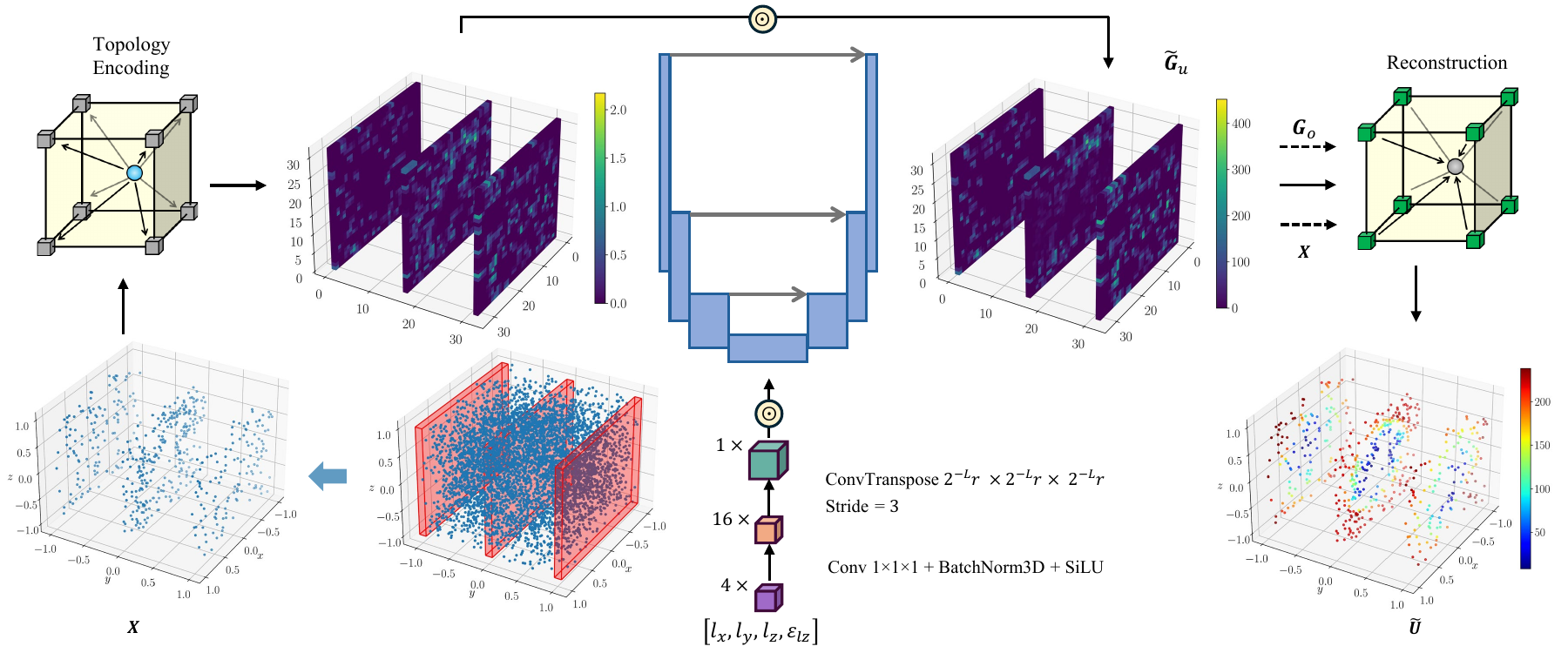}
        \caption{}
        \label{fig flowchart-b}
    \end{subfigure}
    
    \caption{\textbf{Framework flowchart:} Our ML model is based on UNets and learns the mapping between encoded topology and encoded response field. \textbf{(a)} The point cloud is converted to a regular grid via topology encoder. The encoded topology $\boldsymbol{G}_o$ is both the input and the mask applied to the output of the UNet. UNet progressively reduces spatial dimensions and increases channel depth through $L$ contraction blocks ($L = 3$ in this figure). The expansion block inversely mirrors the structure of the contraction block to gradually transform the bottleneck into the output.
     The approximated encoded response $\widetilde{\boldsymbol{G}}_u$ via UNet is then used to reconstruct the responses at the points.
     \textbf{(b)} Our framework naturally extends to 3D. For better visualization, the 3D point clouds are shown in $y \in  [-1.05 , -0.95] \cup [-0.05 , 0.05] \cup [0.95 , 1.05]  $ slices and only the first, middle and last slices of $\boldsymbol{G}_o$ and $\widetilde{\boldsymbol{G}}_u$ along $y$ direction are shown. Extra inputs in 3D cases are treated as channels (e.g., 4 channels in the Solid problem in \Cref{sec results}) and processed through convolutional layers to match the dimensions of the UNet bottleneck; these processed features are then multiplied with the bottleneck.
}
    \label{fig flowchart}
\end{figure}

We design a customized UNet\cite{ronneberger2015u} to map the encoded topology and optional additional inputs to the encoded response (i.e.,  $\boldsymbol{G}_o \mapsto \widetilde{\boldsymbol{G}}_u$), see \Cref{fig flowchart}. Our network has contraction and expansion paths along with a residual connection that leverages the structure of our encoding to improve inference. 

The contraction path extracts hierarchical features by gradually compressing the input spatial dimensions over $L$ levels ($r^2 \rightarrow (r/2)^2 \rightarrow \dots \rightarrow (r/2^L)^2$), while increasing the channel depth ($c_{in} \rightarrow c \rightarrow \dots \rightarrow 2^Lc$). Each block incorporates 2D convolutions with kernel size $k=3$, followed by batch normalization and $SiLU$ activation function for better stability and convergence. Downsampling between consecutive blocks is performed using $2 \times 2$ max-pooling to capture broader contextual features.

The expansion path mirrors the structure of the contraction path, progressively expanding the spatial dimensions to eventually match the original resolution. In the expansion path, upsampling across blocks is handled by convolution coupled with PixelShuffle (aka sub-pixel convolution introduced in \cite{shi2016real}). Pixelshuffle enables detailed spatial recovery by rearranging elements from the depth channel to spatial dimensions based on the upsampling factor $2$, reducing checkerboard artifacts often encountered in transposed convolutions. Furthermore, skip connections in UNet combine low-level details obtained through contraction with high-level semantics provided by expansion via concatenating corresponding feature maps. This approach reduces information loss and preserves fine details. 

To enforce consistency with the encoded topology, we multiply the output of the final layer by the input $\boldsymbol{G}_o$ in an element-wise manner. This operation explicitly zeros out tensor elements representing absent features\footnote{Features whose values are zero since there is no point near them.}, effectively concentrating the model's learning on relevant regions. In other words, UNet is expected to learn the factor $\boldsymbol{G}_u / {\boldsymbol{G}}_o$ which approximates the encoded response $\widetilde{\boldsymbol{G}}_u$ after multiplication with $\boldsymbol{G}_o$.


All the operations in our network have direct higher-dimensional counterparts and so our UNet can be naturally used in more than 2D, see \Cref{fig flowchart-b}. In this figure, we show slices of the original point cloud and its grid representations for better visualization. For additional inputs beyond the point cloud, we can either interpolate them to match the grid resolution $r$ and pass them as extra input channels or process them through convolutional layers to align dimensions at the bottleneck $\dfrac{r}{2^L}$ . In this paper, we adopt the former approach in 2D and the latter approach in 3D since it provides memory and cost gains in 3D problems. 

\subsubsection{Training}\label{subsubsec training}
The only trainable component of our framework is UNet, for which we utilize PyTorch's automatic mixed precision (AMP). AMP leverages $16$-bit floating-point operations during forward passes and gradient scaling during backpropagation, thereby enhancing training efficiency and significantly reducing memory usage.

\begin{figure}[h]
    \centering
    \includegraphics[width=0.99\textwidth]{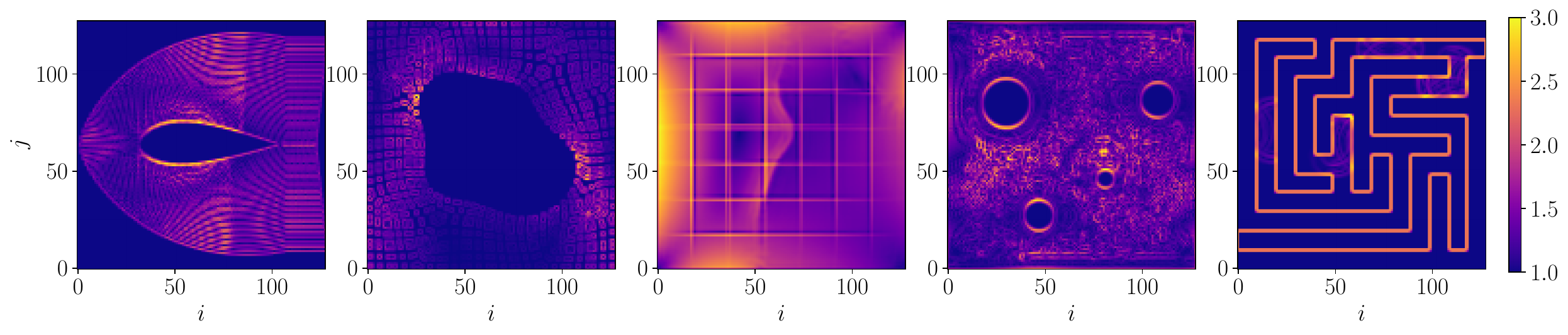} 
    \caption{\textbf{Edge-aware weights:} The weights (i.e., $\boldsymbol{\Gamma}$ in \Cref{eq weight}) associated with representative samples of resolution $r = 128$ are shown. Larger values indicate sharper gradients on which the training should be focused.}
    \label{fig gamma}
\end{figure}

For 2D cases, we define an edge-weighted relative $L_2$ loss to softly localize training near regions with sharp spatial gradients. Given horizontal and vertical Sobel kernels defined as\cite{sobel1990isotropic}:
\begin{equation}
    \boldsymbol{S}_x = 
    \begin{bmatrix}
    -1 & 0 & +1 \\
    -2 & 0 & +2 \\
    -1 & 0 & +1
    \end{bmatrix}, \quad 
    \boldsymbol{S}_y = 
    \begin{bmatrix}
    -1 & -2 & -1 \\
    0 & 0 & 0 \\
    +1 & +2 & +1
    \end{bmatrix},
    \label{eq sobel}
\end{equation}
we compute the spatial gradients of the target encoded response $\boldsymbol{G}_u$, and subsequently calculate the gradient magnitude as follows:
\begin{subequations}
    \begin{align}
    &\boldsymbol{G}_x = \boldsymbol{G}_u * \boldsymbol{S}_x
    , \quad \boldsymbol{G}_y = \boldsymbol{G}_u * \boldsymbol{S}_y\\
    &||\nabla \boldsymbol{G}|| = \sqrt{\boldsymbol{G}_x^2+\boldsymbol{G}_y^2},
    \label{eq sobel2}
    \end{align}
\end{subequations}
where $*$ denotes convolution operation and $||\nabla \boldsymbol{G}||$ is the gradient magnitude based on which we define the edge-aware weights:
\begin{equation}
    \boldsymbol{\Gamma}= 1 + 2\frac{||\nabla \boldsymbol{G}||}{\max{||\nabla \boldsymbol{G}||}}.
    \label{eq weight}
\end{equation}
\Cref{eq weight} assigns spatially varying weights to the domain where regions with minimal gradient magnitude (i.e., least critical regions) receive the lowest weight of $1$, and regions with maximal gradient magnitude (i.e., most critical regions) receive the highest weight of $3$. \Cref{fig gamma} shows instances of $\Gamma$ in the 2D studied examples. Accordingly, we amplify the errors near such edges in the loss function, that is:
\begin{equation}
    \mathcal{L} =\frac{1}{n_b} \sum^{n_b}_{m=1}\frac{||\boldsymbol\Gamma \odot(\widetilde{\boldsymbol{G}}_u - \boldsymbol{G}_u )||_2}{\max(||\boldsymbol{G}_u ||_2 , 10^{-6})},
    \label{eq loss2d}
\end{equation}
where $n_b$ denotes the number of batches. We use a batch size of $M = 10$ for 2D examples and minimize the loss function using AdamW optimizer with the weight decay $2 \times 10^{-2}$. Unless otherwise stated, the initial learning rate is $10^{-3}$, scheduled to reduce by $25\%$ if $\mathcal{L}$ does not improve over $40$ epochs (i.e., its relative change is below 0.01\%). 

We observed no advantage in employing the edge-aware loss for the 3D example where we resort to much coarser grids (e.g., $r \in \{32,64\}$) that yield relatively dense encoded representations with less sharply localized gradients. Therefore, for the 3D problems in \Cref{sec results}, we adopt the standard relative $L_2$ norm as the loss function:
\begin{equation}
    \mathcal{L} =\frac{1}{n_b} \sum^{n_b}_{m=1}\frac{||\widetilde{\boldsymbol{G}}_u - \boldsymbol{G}_u ||_2}{\max(||\boldsymbol{G}_u ||_2 , 10^{-6})}.
    \label{eq loss3d}
\end{equation}
Due to memory limitations, we use a smaller batch size $M=5$ to train the 3D model. All other training hyperparameters are similar to the 2D cases.


\subsection{Error Analysis} \label{subsec error}
The performance of our framework relies on the encoding and reconstruction quality. Therefore, we analyze the error arising from the assumption of local linearity within each cell. For clarity, we focus exclusively on the interpolation scheme in \Cref{eq rec_u}. Nonetheless, topology encoding shares similar error characteristics as it employs the same partitioning method intrinsic to bilinear interpolation.

To facilitate the analysis, we assume the point $(x,y) = (\boldsymbol{x}(m,n) , \boldsymbol{y}(m,n))$ is enclosed in a cell such that  $x_{\si} < x < x_{\si+1}$ and $y_{\sj} < y <y_{\sj+1}$), where:
\begin{equation}
   x_{\si} = h\boldsymbol{i}_{x}^{L}(m , n), \quad x_{\si+1}= h\boldsymbol{i}_{x}^{R}(m , n), \quad y_{\sj}= h\boldsymbol{i}_{y}^{B}(m , n), \quad y_{\sj+1}=h\boldsymbol{i}_{y}^{R}(m , n)~~ \si , \sj \in \{0,...,r-2\}.
\end{equation}
Simliarly, $\boldsymbol{H}(m, \boldsymbol{i}_{x}^{L}(m , n) , \boldsymbol{i}_{y}^B(m , n))$ could be equivalently written as $H_{\mathrm{i},\mathrm{j}}$ for this point.
For notational simplicity, the superscripts of $\Delta x^R$ and $\Delta y ^B$ are dropped hereafter. Therefore, we can  rewrite \Cref{eq rec_u} for point $(x,y)$ as:
\begin{equation} \label{eq u_interp}
    \widehat U(x ,y) = (1- \Delta x)(1- \Delta y) H_{\si , \sj}  +  \Delta x(1- \Delta y)  H_{\si+1 , \sj}+
     (1- \Delta x)\Delta y H_{\si , \sj+1} + \Delta x \Delta y H_{\si+1 , \sj+1} 
\end{equation}

We consider three scenarios in our study, where the response field $H$ has different levels of smoothness in the domain $[-1,1]^2$: $(1) H \in C^2$, i.e., $H$ is twice continuously differentiable; $(2) H \in C^1 \setminus C^2$, i.e., $H$ is once but not twice continuously differentiable; and $(3) H \in C^0 \setminus C^1$, i.e., $H$ is continuous but not continuously differentiable.

\noindent\textbf{(1) If} $\boldsymbol{H \in C^2}$:

We use second-order Taylor expansion around $(x_\si , y_\sj)$ to approximate $H$:
\begin{align} \label{eq h_taylor_c2}
    H(x, y) &= H_{\si,\sj} + (x-x_\si)H_x + (y-y_\sj)H_y  \notag \\
    &+ \frac{1}{2}\parens{(x-x_\si)^2 H_{xx} + 2(x-x_\si)(y-y_\sj)H_{xy} + (y-y_\sj)^2 H_{yy}} + O(h^3) \notag \\
    &=  H_{\si,\sj} + \Delta x h H_x + \Delta y h H_y + \frac{h^2}{2}\parens{\Delta x^2 H_{xx} + 2\Delta x\Delta y H_{xy} + \Delta y^2 H_{yy}} + O(h^3)
\end{align}
We substitute the second-order Taylor expansion approximations in \Cref{eq u_interp}:
\begin{align} \label{eq u_interp_c2}
    \widehat U(x ,y)  &\approx  (1- \Delta x)(1- \Delta y) H_{\si , \sj}  +  \Delta x(1- \Delta y)  (H_{\si,\sj} + hH_x + \frac{1}{2}h^2H_{xx}) \\ \notag
     &+ 
     (1- \Delta x)\Delta y (H_{\si,\sj} + hH_y + \frac{1}{2}h^2H_{yy}) + \Delta x \Delta y(H_{\si,\sj} + hH_x + hH_y + \frac{h^2}{2}(H_{xx} + 2H_{xy} + H_{yy})) \\ \notag
     &=  H_{\si,\sj} + \Delta x h H_x + \Delta y h H_y + \frac{h^2}{2}\parens{\Delta x H_{xx} + 2\Delta x\Delta y H_{xy} + \Delta y H_{yy}}.
\end{align}
By subtracting $\widehat U(x,y)$ in \Cref{eq u_interp_c2} from $H(x,y)$ in \Cref{eq h_taylor_c2} we get the following:
\begin{equation}
    H(x,y) - \widehat{U}(x,y) = \frac{h^2}{2}\Delta x(1-\Delta x)H_{xx} + \frac{h^2}{2}\Delta y(1-\Delta y)H_{yy} + O(h^3)
\end{equation}
Since $\Delta x(1-\Delta x)\leq \frac{1}{4}$ and $\Delta y(1-\Delta y)\leq \frac{1}{4}$, we can bound the error as:
\begin{equation}
|H(x,y) - \widehat{U}(x,y)| \leq \frac{1}{8} \max_{(x,y)}\parens{ |H_{xx}| + |H_{yy}|}h^2
\end{equation}
Under the assumption that the second derivatives of \( H \) are bounded, the interpolation error is given by:
\begin{equation}
|H(x,y) - \widehat{U}(x,y)| = O(h^2) = O\parens{(\sfrac{2}{r-1})^2} \approx O\parens{\sfrac{1}{(r-1)^2}}.
\end{equation}

\noindent\textbf{(2) If }$ \boldsymbol{\ H \in C^1  \backslash C^2}:$

Here, the second derivatives may not exist or may be discontinuous. Therefore, we employ the first-order Taylor expansion around the point $(x_{\si}, y_{\sj})$:
\begin{align} \label{eq h_taylor_c1}
H(x,y) &= H_{\si,\sj} + (x - x_{\si})H_x + (y - y_{\sj})H_y + O(h^{1+\alpha}) \notag \\
&= H_{\si,\sj} + \Delta x h H_x + \Delta y h H_y + O(h^{1+\alpha})
\end{align}
where $\alpha = 1$ when the first derivatives are Lipschitz continuous; $ 0<\alpha < 1$ holds when the first derivatives are only Hölder continuous, and $\alpha = 0$ if first derivatives exist but are not Hölder continuous.
We substitute the first-order Taylor expansion approximations into \Cref{eq u_interp} and simplify:
\begin{align}
\widehat U(x,y) &\approx (1-\Delta x)(1-\Delta y)H_{\si,\sj} + \Delta x(1-\Delta y)(H_{\si,\sj} + hH_x) \notag \\
&+ (1-\Delta x)\Delta y(H_{\si,\sj} + hH_y) + \Delta x\Delta y(H_{\si,\sj} + hH_x + hH_y) \notag \\
&= H_{\si,\sj} + \Delta x hH_x + \Delta y hH_y.
\end{align}
Thus, the leading-order error term becomes:
\begin{equation}
|H(x,y) - \widehat{U}(x,y)| = O(h^{1+\alpha}) = O\parens{(\sfrac{2}{r-1})^{1+\alpha}} \approx O\parens{\sfrac{1}{(r-1)^{1+\alpha}}},
\end{equation}
where $\alpha \in [0,1]$ depends on the continuity of the first derivatives of $H$.

\noindent\textbf{(3) If }$ \boldsymbol{\ H \in C^0  \backslash C^1}:$

In this case, we cannot perform a Taylor expansion. Without differentiability, the interpolation error depends solely on continuity conditions, generally exhibiting at most first-order accuracy $O(h)$, often lower:
\begin{equation} \label{eq err discont}
|H(x,y) - \widehat{U}(x,y)| = O(h^{\gamma}) = O\parens{(\sfrac{2}{r-1})^{\gamma}} \approx O\parens{\sfrac{1}{(r-1)^{\gamma}}}, \quad \gamma \in[0,1].
\end{equation}
More specifically, if $H$ is Hölder continuous (i.e., $\gamma \in (0,1)$), the interpolation accuracy is sublinear. If it lacks Hölder continuity, the convergence rate becomes extremely slow and it barely improves with refinement $O(1)$.

An error bound of $O(h^\beta)$ with $\beta \in[0,2]$ provides a guideline for balancing accuracy against memory and computational cost in our grid-based scheme. Since the worst-case interpolation error scales as $h^\beta$, halving the grid spacing reduces the maximum error by approximately a factor of $2^\beta$, so higher values of $\beta$ (corresponding to smoother functions) enable coarser grids to meet a target error tolerance. Conversely, one can preselect $h \approx (\epsilon/\eta)^\frac{1}{\beta}$ to guarantee an error below $\epsilon$, where $\eta$ encapsulates all problem-specific factors that do not depend on $h$. In practice however, the smoothness level might vary spatially and $\eta$ parameter might be unknown. Nevertheless, $O(h^\beta)$ bound is still useful as it describes the interpolation error behavior and helps explain phenomena such as error stagnation in non-smooth regions and substantial error reduction from further grid refinement in smooth regions.

\subsection{Time Complexity Analysis} \label{subsec complexity}
Understanding the time complexity of our framework is crucial for assessing its scalability and efficiency, especially when dealing with large-scale point cloud data. The framework consists of two main components: an encoding scheme that converts point clouds and their corresponding responses into grid representations and vice versa,  and a UNet-based architecture that maps these structured grids. Analyzing the computational cost of each component allows us to identify potential bottlenecks and determine how the framework scales with key parameters such as the number of input points ($N$), the grid resolution ($r$), and the depth of the network ($L$). In the following sections, we provide a detailed complexity analysis of both the encoding process and the UNet model, outlining how different operations contribute to the overall computational cost.

\subsubsection{Encoding}
As described in \Cref{subsec encoding}, we convert point clouds and their corresponding responses to grid representations and recover values at the original point cloud using three key functions - topology encoding, response encoding, and reconstruction. We study the performance of each function in terms of time complexity for a single point cloud (i.e., $M = 1$). 

\textbf{Topology Encoding}: 
Mapping the scaled coordinates to grid index space involves a few arithmetic operations (addition, multiplication, division) that incur a cost of $O(N)$ for each point cloud sample. Computing neighboring gird points is also constant-time per point, i.e.,  $O(N)$, due to the simple floor and addition operations. Calculating bilinear interpolation weights using subtractions and multiplications approximately costs $O(N)$, and scattering the spatial contributions of each point to the four grid vertices is $O(4N)$. Therefore, the topology encoding process scales linearly with the number of points in the point cloud, yielding a cost of $O(N)$ for a single sample.

\textbf{Response Encoding}: 
The steps in response encoding closely follow those of topology encoding, except that the goal is to accumulate response values rather than spatial contributions from each point in the point cloud onto the grid. Thus, the final step differs slightly with extra multiplications (see \Cref{eq response encoding}), but still maintains an overall computational cost of approximately $O(N)$.

\textbf{Reconstruction}:
Even though many steps in reconstruction (coordinate transformation, neighboring grid points identification, and obtaining interpolation weights) are similar to the topology and response encoding whose cost is $O(N)$, reconstruction involves \textit{reading} from the grid, rather than \textit{writing} to it, which requires an additional step involving element-wise division to obtain the normalized grid $\boldsymbol{H}$ in \Cref{eq rec_u}. Since the grid size is $r \times r$, the associated cost is $O(r^2)$. Consequently, the overall cost of reconstruction can be approximated as $O(N+r^2)$ 

Combining the costs of topology encoding, response encoding, and reconstruction, the total computational cost of the encoding process is approximately $\mathcal{C}_{encoding} = O(3N + r^2)$. This result confirms that the encoding scales linearly with the number of points in the point cloud, with an additional quadratic term due to the reconstruction. 

\subsubsection{UNet}
In the U-shaped architecture, the computational cost is mostly dominated by convolution layers as they require a large number of multiply-accumulate (MAC) operations across all pixels and channels. Since each convolution in the network involves a kernel of fixed size ($k=3$) and operates over feature maps that change in spatial resolution and channel depth at different stages of the model, analyzing the time complexity of the UNet part of our model largely reduces to counting the number of convolutions and their corresponding input and output dimensions. Thus, in the following analysis, we focus on the cost introduced by convolutions in each block (initial, contraction, expansion, and final) to derive the total computational complexity.

The initial block of the U-shaped architecture is composed of two successive convolutions that maintain the spatial dimensions (i.e., resolution $r$) while increasing the number of channels to the base hidden channel count $c$. Given the input size $(M, c_{in}, r, r)$, the per-sample computational cost of the first and second convolutions are written as:
\begin{equation}
    \mathcal{C}_{0} = \mathcal{C}_{0,1} + \mathcal{C}_{0,2} =  O(r^2 k^2 cc_{in}) +  O(r^2 k^2 c^2) = O\parens{r^2 k^2c (c+c_{in})}
    \label{eq init_block}
\end{equation}
The output of the initial block is a tensor of size  $(M, c, r, r)$ which is passed to the contraction path. Each block $i = 1,..., L$ in the contraction path first applies a $2 \times 2$ max pooling which reduces the spatial resolution by a factor of $2$, then applies two convolutions for doubling the channel count. Assuming negligible cost for max-pooling layers, the first and second convolutions' costs are as follows:
\begin{subequations}
    \begin{align}
        &\mathcal{C}_{c,1}^i = O\parens{\frac{r^2}{4^i} k^2 \parens{2^{i-1}c} \parens{2^{i}c}} = O\parens{\frac{1}{2} r^2 k^2 c^2}, \\
        &\mathcal{C}_{c,2}^i = O\parens{\frac{r^2}{4^i} k^2 \parens{2^{i}c}^2} = O\parens{r^2 k^2 c^2}.
    \end{align}
    \label{eq cblock}
\end{subequations}
Therefore, the total cost of the contraction path scales linearly with the number of blocks:
\begin{equation}
    \mathcal{C}_c = \sum_{i=1}^L \mathcal{C}_{c}^i = \sum_{i=1}^L \mathcal{C}_{c,1}^i + \mathcal{C}_{c,2}^i =  O\parens{\frac{3L}{2}r^2 k^2 c^2}
    \label{eq cpatc}
\end{equation}

Each expansion block mirrors a corresponding contraction block. For each expansion block, we let the matching index be $i = L -j+1$ for $j = 1,..,L$.  The expansion path starts with convolving contracted input of size  $(M, 2^Lh , \frac{r}{2^L}, \frac{r}{2^L})$ with an upscale factor of $\nu=2$ and a target output channel $2^{i-1}h$, which approximately costs:
\begin{equation}
    \mathcal{C}_{e,1}^i = O\parens{\frac{r^2}{4^i}k^2(2^ic)(\nu^2 2^{i-1}c)} = O(2r^2k^2c^2).
    \label{eq eblock1}
\end{equation}
This convolution is followed by a PixelShuffle operation, which rearranges the resulting tensor from shape $(M, 4 \times 2^{i-1}c, \frac{r}{2^i}, \frac{r}{2^i})$ to $(M, 2^{i-1}c,  \frac{r}{2^{i-1}}, \frac{r}{2^{i-1}})$ with negligible computational cost compared to convolution. The resulting tensor is then concatenated with the output from the corresponding contracting block with the relatively small cost of $O\parens{(\frac{r}{2^{i-1}})^2}$.
The concatenated tensor is then processed by two convolutions per block to gradually reduce the number of channels. The cost of the subsequent convolutions are:
\begin{subequations}
    \begin{align}
        &\mathcal{C}_{e,2}^i = O\parens{\frac{r^2}{4^{i-1}} k^2 (2\times 2^{i-1}c)(2^{i-1}c)} = O(2r^2k^2c^2), \\
        &\mathcal{C}_{e,3}^i = O\parens{\frac{r^2}{4^{i-1}} k^2 (2^{i-1}c)^2} = O\parens{r^2 k^2 c^2}.
    \end{align}
    \label{eq eblock23}
\end{subequations}
Hence, the total cost of the expansion path is approximated as:
\begin{equation}
    \mathcal{C}_e = \sum_{i=1}^L \mathcal{C}_{e}^i = \sum_{i=1}^L \mathcal{C}_{e,1}^i + \mathcal{C}_{e,2}^i + \mathcal{C}_{e,3}^i =  O\parens{5Lr^2 k^2 c^2}.
    \label{eq epatc}
\end{equation}
In the final layer, a $1 \times 1$ convolution is applied to reduce the number of channels from $h$ to the output channel $c_{out}$ (e.g., $c_{out} = 1$ in single-output problems). Thus, the final layer's cost is $\mathcal{C}_{f} =  O(r^2 cc_{out})$.

Treating $k$ as a constant value and leaving out all constant terms, we can simplify and write the total cost of the 2D UNet as:
\begin{align}
    \mathcal{C}_{UNet} &=  \mathcal{C}_{0} + \mathcal{C}_{c} + \mathcal{C}_{e} + \mathcal{C}_{f} \notag \\
    &\approx O\parens{r^2 c \parens{c+c_{in}}} +  O\parens{L r^2 c^2} + O\parens{L r^2 c^2} + O\parens{r^2 cc_{out}} \notag \\
    &= O\parens{r^2 \parens{cc_{in} + (1+2L)c^2 + cc_{out}}}.
    \label{eq unet_cost}
\end{align}
\Cref{eq unet_cost} implies that for fixed values of hyperparameters $c$ and $L$, and problem-specific parameters $c_{in}, c_{out}$, computational complexity increases quadratically with resolution $r$.

The total computational cost of our framework combines the encoding and the UNet cost as follows:
\begin{align}
    \mathcal{C}_{total} &=  \mathcal{C}_{encoding} + \mathcal{C}_{UNet}  \notag \\
    &= O\parens{3N + r^2} +  O\parens{r^2 \parens{cc_{in} + (1+2L)c^2 + cc_{out}}} \notag \\
    &\approx O\parens{N + r^2 \parens{1 + (c_{in}+c_{out})c + (1+2L)c^2 }}.
    \label{eq total_cost}
\end{align}
This expression highlights that the encoding process scales linearly with the number of points $N$ while the UNet introduces a quadratic dependence on resolution $r$. Consequently, for high resolutions the computational cost is dominated by the quadratic terms particularly those associated with the UNet’s convolutional operations. In 3D cases, where the grid resolution extends to $r \times r \times r$, the computational complexity of both encoding and UNet further increases as the dependence on $r$ becomes cubic.

    \section{Results and Discussions} \label{sec results}

We consider five canonical 2D examples and one 3D example simulated with different types of mesh to evaluate our framework (E-UNet). The 2D examples are used to compare our model against baseline models FNO(-interp)\cite{li2020fourier}, GeoFNO\cite{JMLR:v24:23-0064}, and GNOT\cite{hao2023gnot}. Additionally, we integrate our encoding scheme with FNO (E-FNO) to demonstrate its effectiveness compared to conventional encoding/interpolation techniques. For the 3D example, we exclude FNO variants from our comparative studies (due to their high memory requirements that exceed our resources) and benchmark our framework against GNOT and GeomDeepONet \cite{he2024geom}. For each problem, we train the models multiple times to ensure the reported values are representative. Detailed statistics regarding these random trials and models' configurations are included in \Cref{app stats} and \Cref{app model config}, respectively. All models are trained on an NVIDIA RTX 4090 GPU.

The benchmark problems are described in \Cref{subsec benchmark descriptions} and the outcomes of our comparative studies are provided in \Cref{subsec comparative studies}. We assess the robustness of our framework to noise and limited training data in  \Cref{subsec sensitivity analysis} and then in \Cref{subsec inverse mapping} we elaborate on how our approach can be used for recovering responses at the missing points in the point cloud given partial observations.

\subsection{Description of the Benchmark Problems} \label{subsec benchmark descriptions}
\noindent\textbf{NACA:}
The inviscid compressible steady transonic flow over multiple airfoils is studied in \cite{JMLR:v24:23-0064} which is governed by the Euler equations in conservative form: 
\begin{subequations}
    \begin{align}
        &\nabla \cdot(\rho \boldsymbol{u}) = 0, \\
        & \nabla \cdot(\rho \boldsymbol{u} \otimes\boldsymbol{u}+p \boldsymbol{I}) = \boldsymbol{0} ,\\
        & \nabla \cdot ((E+p)\boldsymbol{u}) =0,
    \end{align}
    \label{eq naca}
\end{subequations}
where $\rho$ is the fluid density, $\boldsymbol{u}$ is the velocity vector, $p$ is the pressure, $E$ is the total energy, and $\otimes$ denotes outer product. No penetration boundary condition is imposed on the airfoils. The far-field flow conditions are: $\rho_\infty = 1 ,~ p_\infty = 1$, and the far-field Mach number is $M_\infty = 0.8$, giving rise to the formation of shock waves in many samples in the dataset. The simulations are performed using C-grid mesh from which the Cartesian coordinates of the nodes can be extracted. The mesh is refined near the airfoil, but not around the shock.  The goal is to predict the Mach number at each node given the point cloud describing the airfoil shape and its angle of attack. We use $n_{train} = 1000$ training samples and $n_{test}=100$ test samples, each sample has $N = 2820$ points.

\noindent\textbf{Hyperelasticity}:
A solid body spanning the $[0,1]^2$ domain with an irregular void at the center is clamped at the bottom edge and subjected to a tensile traction $\boldsymbol{\mathrm{t}}$ on the top boundary.  The material follows the incompressible Rivlin-Saunders model with energy density:
\begin{equation}
    \begin{aligned}
        &w(\boldsymbol{\mathrm{E}}) = C_1(I_1 - 3) + C_2(I_2 - 3), \\ 
        &I_1 = \Tr{(\boldsymbol{\mathrm{C}})} , \quad
        I_2 = \frac{1}{2}[ \Tr{(\boldsymbol{\mathrm{C}})^2 - \Tr((\boldsymbol{\mathrm{C}}^2)}], 
    \end{aligned}
    \label{eq energy helas}
\end{equation}
where $\boldsymbol{\mathrm{C}} = 2\boldsymbol{\mathrm{E}} + \boldsymbol{I}$ is the Right Cauchy-Green tensor, $\boldsymbol{\mathrm{E}}$ is the Green-Lagrange strain tensor and the energy density function parameters are $C_1 = 1.863 \times 10^5$ and $C_2 =9.79 \times 10^3$. The solid body is in the equilibrium condition:
\begin{equation}
    \begin{aligned}
        \nabla \cdot \parens{\frac{\partial w}{\partial \boldsymbol{\mathrm{E}}}} = \nabla \cdot \boldsymbol{\sigma} = \boldsymbol{0}. 
    \end{aligned}
    \label{eq helas}
\end{equation}
The domain is meshed with quadrilateral elements, resulting in $N = 972$ nodes whose collection is treated as a point cloud. The traction is uniformly discretized into $42$ points $\boldsymbol{\mathrm{t}} \in [0.2,0.4]^{42}$. The goal is to map the point cloud and the discretized traction to the effective stress field defined in \cite{JMLR:v24:23-0064}. In grid-based models, we linearly interpolate $\boldsymbol{\mathrm{t}}$ and expand it to match the grid resolution ($r$) and feed it as an additional input channel to the model. We use $n_{train} = 1000$ training samples and $n_{test}=100$ test samples from the dataset.

\noindent\textbf{Darcy}: The flow through unit square porous media with the permeability field $\kappa(\mathrm{x},\mathrm{y})$ and a constant source term of $1$ is governed by the following PDE:
\begin{equation}
    \begin{aligned}
       -\nabla \cdot (\kappa \nabla p) = 1,
    \end{aligned}
    \label{eq darcy}
\end{equation}
where $p$ is the pressure which is set to $0$ on the boundaries. The dataset containing the solution $p$ corresponding to different permeability fields is provided in \cite{li2020fourier}. The simulations are performed on a $241 \times 241$ regular grid mesh which we subsample every other point for our model and for other models we interpolate it to match the chosen resolution ($r$) for our encoding. We use $n_{train} = 1000$ training samples and $n_{test}=100$ test samples from this dataset. We highlight that even though the original data is on a regular mesh, the nodes do not necessarily coincide with our grid vertices and hence no specific advantage or disadvantage is given to our approach while other methods such as FNO variants directly benefit from this setup.

\noindent\textbf{Circles}:
Four circles with different radii are randomly placed in the square domain $[0,8]^2$. The left side is the flow inlet for which the velocity $\boldsymbol{u}_{inlet} = [\mathrm{y}(8-\mathrm{y})/16 , 0]^\top$ is prescribed. The right side is the outlet with the prescribed pressure $p_{outlet} = 0$. No-slip condition is imposed on all domain walls. The steady flow over the four circles is governed by the Navier-Stokes equations in 2D: 
\begin{subequations}
    \begin{align}
        &(\boldsymbol{u} \cdot \nabla ) \boldsymbol{u} -\frac{1}{Re}\nabla^2 \boldsymbol{u} + \nabla p= \boldsymbol{0}, \\
        & \nabla \cdot \boldsymbol{u} =0,
    \end{align}
    \label{eq circles}
\end{subequations}
where $\boldsymbol{u}$ is the velocity vector, $p$ is the pressure, and $Re$  matches the Reynolds number used in \cite{hao2023gnot}. The goal is to predict $[\boldsymbol{u} , p]^\top$ for all points in the point clouds associated with different layouts of the domain. Triangular elements are used for meshing the domain where the mesh is refined near the walls. 
The dataset is provided in \cite{hao2023gnot} and we use $n_{train} = 1000$ for training and $n_{test} = 100$ for testing. The minimum and maximum number of points in the samples are $N_{min} = 8735$ and $N_{max} = 11446$, respectively. For GeoFNO which cannot handle point clouds of different sizes, all point clouds are downsampled to have $N_{min}$ points. 

\noindent\textbf{Maze}:
Pressure waves initiate from point sources and propagate through the domain $[-1,1]^2$ consisting of low-density maze paths and orders of magnitude higher-density maze walls. The system governing wave propagation in heterogeneous media is:
\begin{subequations}
    \begin{align}
        &\frac{\partial p}{\partial t} + K \nabla \cdot \boldsymbol{u} = 0, \\
        &\frac{\partial \boldsymbol{u}}{\partial t} + \frac{1}{\rho}\nabla p  = \boldsymbol{0},
    \end{align}
    \label{eq maze}
\end{subequations}
where $K$ is the bulk modulus set to $4$, $\boldsymbol{u}$ and $p$ are the velocity and pressure, respectively, and $\rho$ is the density which is $10^6$ for the walls and $3$ for the paths. 
The original data \cite{ohana2024well,mandli2016clawpack} is arranged in a regular $256 \times 256$ spatial grid in the time interval $[0 , 4]$ with $\Delta t = \frac{2}{201}$ increments. The goal is to obtain the pressure field $p$ at $10 \Delta t$ from the maze layout and the initial pressure rings. Since the waves propagate through paths only, we discard the walls in all samples and treat the remaining points as the point clouds used for training all models except for FNO. Each point cloud in the resulting dataset contains between $N_{min} = 6228$ and $N_{max} = 7399$ points. For GeoFNO, we randomly downsample all point clouds to $N_{min}$ points, and for FNO, we downsample the original grid based on the chosen resolution. We use $n_{train} = 1600$ and $n_{test} = 200$ samples for training and testing, respectively. In addition, the initial learning rate is set to $10^{-4}$ for this problem.

\noindent\textbf{3D solid}:
A $l_x \times l_y \times l_z$ cuboid solid body containing a randomly oriented ellipsoid void is simply supported at $z=0$ and subjected to a tensile strain $\varepsilon_{lz} \in [0.10, 0.15]$ on the opposite face. An elasto-plastic material with linear isotropic hardening is considered with Young's modulus of $209$ GPa, Poisson's ratio of $0.3$, yield stress of $235$ MPa, and a hardening modulus of $800$ MPa. The geometries were meshed by quadratic tetrahedral elements in \cite{he2024geom}, resulting in an average node count of $45649$. The nodes are randomly downsampled to $N = 5000$ points per point cloud. The goal is to map the point clouds and the tensile strain $\varepsilon_{lz}$ to their corresponding von Mises stress.

Since the domain size varies significantly in this dataset, i.e, $l_x , l_y , l_z \in [2.2 , 14.9]$, we transform each cuboid domain to the reference cuboid $[-1 , 1]^3$ to fully leverage the capacity of our encoding scheme and pass $[l_x , l_y , l_z]$ along with the tensile strain $\varepsilon_{lz}$ as additional inputs to the UNet's bottleneck as shown in \Cref{fig flowchart-b}. We use $n_{train} = 2000$ to train our model, GeomDeepONet, and GNOT.  $n_{test} = 500$ samples are used to test all the models.

\subsection{Outcomes of Comparative Studies} \label{subsec comparative studies}
We perform extensive quantitative and qualitative studies to benchmark E-UNet in 2D (\Cref{subsubsec 2d}) and 3D (\Cref{subsubsec 3d}) setups across the examples introduced in \Cref{subsec benchmark descriptions}. We compare the relative $L_2$ error on the test sets, training costs, and the visual fidelity of the predicted fields. We further examine how resolution $r$ influences the models' training and predictive performance.

\subsubsection{2D Cases} \label{subsubsec 2d}
\begin{figure}[!t]
    \centering
    \includegraphics[width=0.99\textwidth]{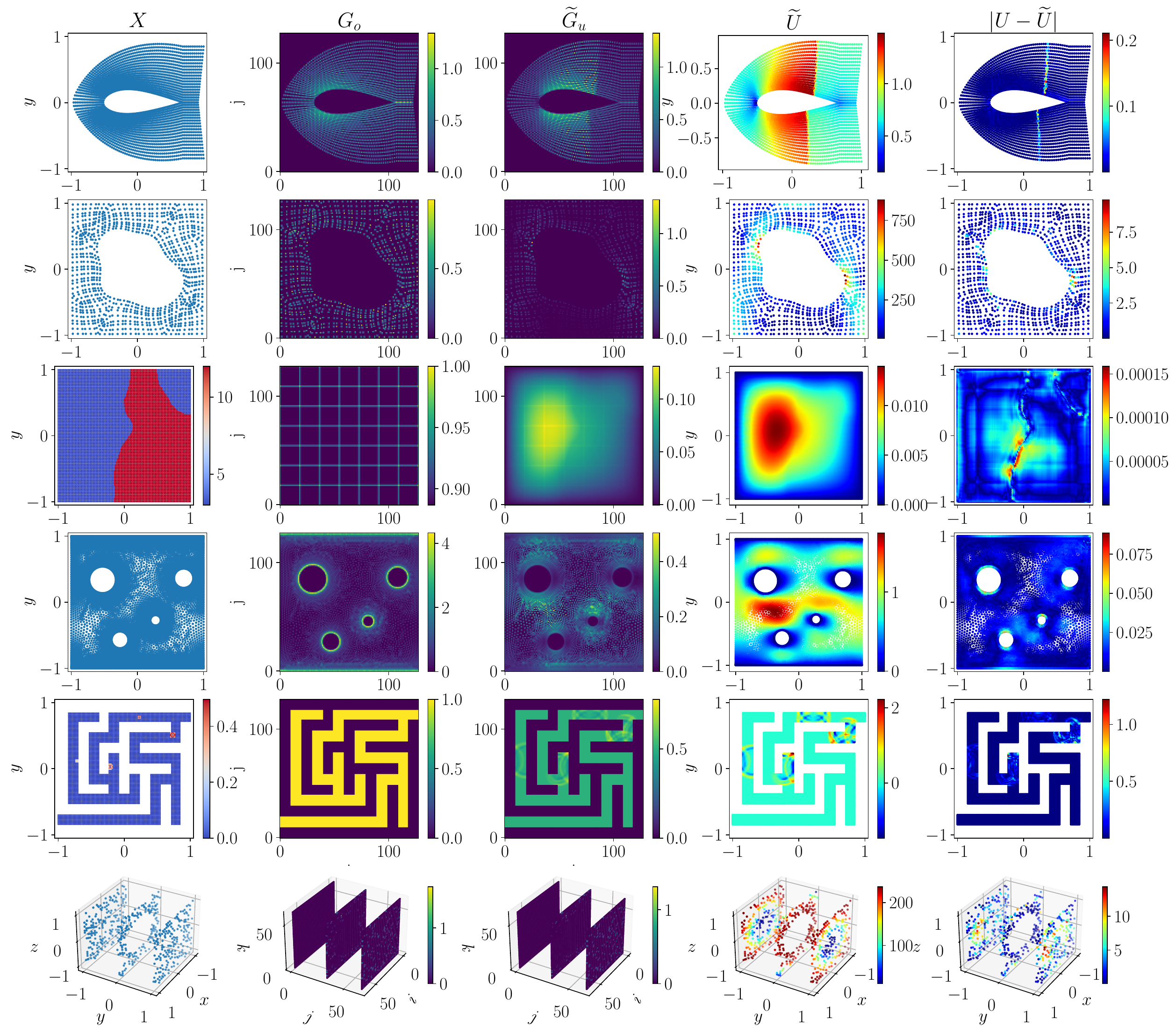} 
    \caption{\textbf{ Visualization of the key components of the proposed E-UNet framework for a representative test sample from each problem at resolution $r=128$:} From left to right the columns indicate input coordinates $\boldsymbol{X}$, encoded topology $\boldsymbol{G}_o$, predicted encoded reponse $\boldsymbol{\widetilde{G}}_u$, reconstructed solution $\boldsymbol{\widetilde{U}}$, and the corresponding absolute error $|\boldsymbol{U} - \boldsymbol{\widetilde{U}}|$. The error maps correspond to the median-performing model across five independent training runs. In Darcy and Maze instances, the permeability and the initial pressure fields are also shown at $\boldsymbol{X}$, respectively. In Circles, the $x$-component of the velocity is considered. For better visualization in the case of the 3D problem (bottom row), results are shown only in the first, last, and middle planes.}
    \label{fig all128}
\end{figure}

We train the 2D models, E-UNet, E-FNO, FNO(-interp) and GeoFNO for $1000$ epochs. GNOT is trained for $500$ epochs based on the recommended configuration in the literature as higher numnber of epochs can potentially result in overfitting. We evaluate the performance of the models using the relative $L_2$ metric, that is:
\begin{equation}
   L_2^{re} = \sqrt{\frac{\sum_{m=1}^{n_{test}}\sum_{n=1}^{N_{m}}(\boldsymbol{U}(m , n) - \boldsymbol{\widetilde{U}}(m , n))^2}{\sum_{m=1}^{n_{test}}\sum_{n=1}^{N_{m}}\boldsymbol{U}(m , n)^2}},
    \label{eq rl2}
\end{equation}
where $\boldsymbol{U}(m , n)$ is the ground truth response corresponding to sample $m$ in the testset at the $n^{th}$ point in the point cloud, $N_m$ is the total point count in the $m^{th}$ sample, and $\boldsymbol{\widetilde{U}}(m , n)$ is the approximated response predicted by the model.
In our framework (E-UNet), $\boldsymbol{\widetilde{G}}_u$ is first approximated through the UNet backbone, then reconstructed into $\boldsymbol{\widetilde{U}}$ via \Cref{eq rec_u} based on the positions $\boldsymbol{X}$ and the encoded topology $\boldsymbol{G}_o$. \Cref{fig all128} displays the aforementioned quantities at resolution $r=128$ for a representative test sample from each problem. The absolute errors associated with the median performance of five independently trained models are shown in the rightmost column of \Cref{fig all128}.

The backbone of FNO(-interp), E-FNO and GeoFNO is the vanilla FNO architecture with $12$ frequency modes and a hidden channel dimension of $32$ across all examples \cite{li2020fourier}. FNO(-interp)'s input is the binary mask of the domain described by the point cloud, marking every grid vertex inside the domain by $1$ and every vertex outside the domain by $0$ (e.g., zero values are assigned to vertices inside the airfoil and far away from it in the NACA example). Its target output is the bilinear interpolation of the field responses onto an $r \times r$ grid with nearest-neighbor fallback outside the convex hull of the scattered points. This hybrid interpolation allows for better reconstruction accuracy along the boundaries.

To study the effect of resolution on models' performance, we train the resolution-based models E-UNet, E-FNO, and FNO(-interp)  at $r \in \{32 , 64 , 128 , 256\}$ and bilinearly interpolate the output back onto the original point cloud.
Whereas E-UNet and E-FNO are trained following the procedure outlined in \Cref{subsubsec training} (i.e., an edge-aware mask is applied to the error map during training, see \Cref{fig gamma}),  FNO(-interp) is trained by minimizing the relative $L_2$ loss using the Adam optimizer, paired with a multi-step learning rate scheduler that reduces the learning rate by $25\%$ five times over $1000$ epochs. 

\begin{figure}[!t]
    \centering
    \includegraphics[width=0.99\textwidth]{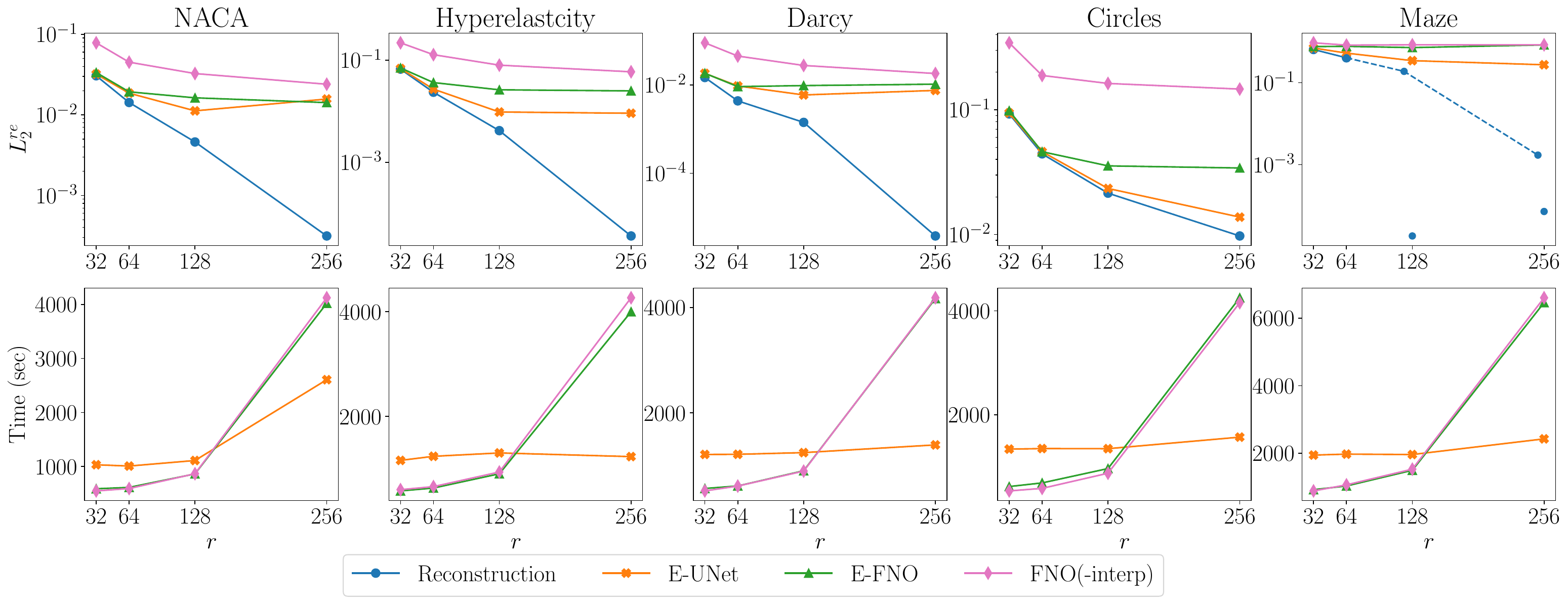} 
    \caption{\textbf{Comparison of accuracy and training time for different resolutions:} Higher-resolution models generally achieve improved accuracy in terms of $L_2^{re}$ at the expense of increased training time. E-UNet offers better scaling with resolution compared to E-FNO and FNO(-interp) and approaches closer to the reconstruction error limit. The errors are calculated based on the velocity magnitude for Circles.}
    \label{fig res}
\end{figure}

The median $L_2^{re}$ and training times from five independent training runs are presented in \Cref{fig res} which reveals distinct error behaviors across different resolutions.
We observe that at lower resolutions (i.e., $r\in \{32 ,64\}$) the reconstruction error arising from bilinear interpolation (i.e., \Cref{eq rl2} based on $\widehat{\boldsymbol{U}}$ instead of $\widetilde{\boldsymbol{U}}$) dominates the total error. However, at higher resolutions (i.e., $r \in \{128 ,256\}$) the reconstruction error becomes negligible and the total error is dominated by the approximation error due to inaccuracies associated with the learning the underlying mapping function. Among the models, E-UNet consistently demonstrates better scalability, achieving lower errors with increasing resolution while maintaining computational efficiency compared to E-FNO and FNO(-interp).
 
Additionally, E-FNO demonstrates better accuracy than FNO(-interp) across all benchmark problems while incurring roughly the same computational cost, despite the architectural similarity between the two models. This improvement is ascribed to two key factors: (1) the input to E-FNO carries richer information content as it encodes topological features beyond the binary representation of the domain used in FNO(-interp) (see \Cref{subsubsec info} for information content analysis); (2) The training process is more targeted in E-FNO, benefiting from the edge-aware weighting of the loss and the post-multiplication with $\boldsymbol{G}_o$ which discards the irrelevant regions of the output and aligns the training more closely with the structural similarity between the encoded geometry and its corresponding response.

A notable exception occurs in the Maze example where generalization error dominates due to the extrapolative nature of the test set. Even though E-UNet outperforms other models in this case, its median training relative $L_2$ error ($3.48E\!-\!02$) is still significantly lower than its test $L_2^{re}$ ($2.72E\!-\!01$). The dashed blue line in the Maze plot in \Cref{fig res} represents reconstruction errors specifically evaluated at resolutions $r \in \{120,250\}$, suggesting that the reconstruction errors reported at $r\in \{128,256\}$ are unrealistically low as they match the original resolution of the data.

We present the summary of comparative studies in \Cref{tab summary}. The median test metric $L_2^{re}$ and training time are reported for the best-performing $r$ for resolution-based models. The comprehensive statistics of the results are provided in \Cref{app stats}. \Cref{tab summary} indicates that E-UNet consistently outperforms other models across most problems, except for Circles. In this multi-output case, GNOT achieves $15\%$ lower $L_2^{re}$ for the velocity magnitude, albeit at a significantly higher computational cost. We attribute the shortcoming of E-UNet in this problem to its simplistic summation of error terms corresponding to each velocity component $u$, $v$ and the pressure $p$, which may not effectively capture their relative importance. We also note from \Cref{fig res} that $L_2^{re}$ for Circles does not appear fully converged with respect to resolution. While further improvements may be possible at higher resolutions, we maintain the same resolution range across all problems for consistency.

\begin{table}[!b]
\centering
\footnotesize
\renewcommand{\arraystretch}{1.1}
\setlength{\tabcolsep}{2pt} 
\begin{tabular}{|c|cc|cc|cc|cc|cc|}
\hline
\textbf{Problem} 
& \multicolumn{2}{c|}{\textbf{E-UNet}} 
& \multicolumn{2}{c|}{\textbf{E-FNO}} 
& \multicolumn{2}{c|}{\textbf{FNO(-interp)}} 
& \multicolumn{2}{c|}{\textbf{GeoFNO}} 
& \multicolumn{2}{c|}{\textbf{GNOT}} \\
& $L_2^{re}$  & Time (sec)
&  $L_2^{re}$ & Time (sec)
& $L_2^{re}$ & Time (sec)
&  $L_2^{re}$ & Time (sec)
&  $L_2^{re}$ & Time (sec)\\
\hline
\textbf{NACA} & \textbf{1.12E-02} & 1112  & 1.42E-02 & 4021 & 2.40E-02 & 4123 & 1.90E-02 & 998 & 1.21E-02 &1835  \\
\textbf{Hyperelasticity} & \textbf{9.11E-03} & 1233  & 2.50E-02 & 3999 & 5.91E-02 & 4266 & 1.97E-02 & 1075 & 1.32E-02 &   1757  \\
\textbf{Darcy} & \textbf{5.94E-03} & 1248 & 9.68E-03 & 904 & 1.82E-02 & 4183 & 9.42E-03 & 1201 & 2.77E-01 & 10049 \\
\textbf{Circles} & 1.38E-02 & 1602 & 3.41E-02 & 4338 & 1.46E-01 & 4155 & 1.02E-01 & 3126 & \textbf{1.18E-02} & 5578 \\
\textbf{Maze}   & \textbf{2.72E-01} & 2426 & 7.15E-01 & 1495 & 8.14E-01 & 6603 & 7.60E-01 & 3660 & 9.941E-01 & 5259 \\
\hline
\end{tabular}
\caption{\textbf{Comparison of the 2D models:} Median $L_2^{re}$ and training time of $5$ training repetitions are reported for the resolution with the most accurate predictions from E-UNet, E-FNO, and FNO(-interp). The errors are calculated based on the velocity magnitude in Circles.}
\label{tab summary}

\end{table}

It is noteworthy to mention that GeoFNO technically reduces to FNO in problems such as Darcy where the data is defined on regular grids. However, GeoFNO still outperforms other FNO variants in Darcy due to its larger padding size ($8$ rather than $4$) which induces stronger artificial periodicity and better accommodates the Fourier transform’s inherent assumption of periodic boundary conditions. 

\begin{figure}[!t]
    \centering
    \includegraphics[width=0.99\textwidth]{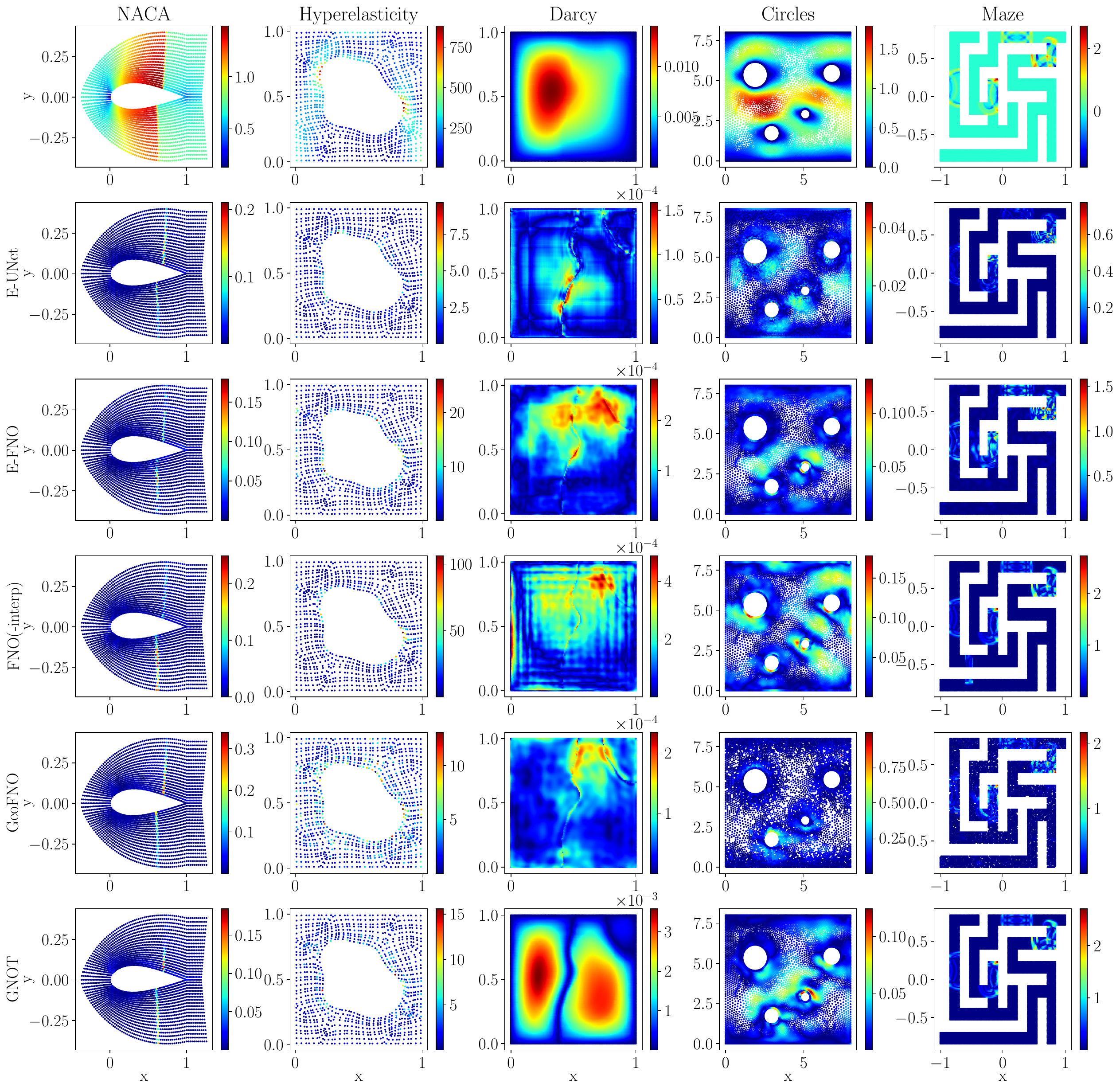} 
    \caption{\textbf{ Visualization of the ground truth and the error maps:} Representative samples from each problem are taken for which the error maps corresponding to the median performing model among $5$ training repetition are shown. All models struggle with sharp gradients seen in NACA and Maze.}
    \label{fig errors}
\end{figure}

To better interpret the quantitative results in \Cref{tab summary}, in \Cref{fig errors} we visualize the corresponding error maps obtained from the median-performing models across $5$ training repetitions for a representative test sample from each problem. We observe that all models struggle to capture the shock in NACA. As discussed in \Cref{subsec error}, We expect this error localization for our encoding strategy due to its direct dependence on the smoothness level of the solution field. Even though E-FNO and FNO(-interp) both rely on bilinear interpolation with similar backbone architecture, FNO(-interp) exhibits more pronounced checkerboard artifacts and higher errors near edges caused by its primitive binary input and non-refined bilinear interpolation. 

While GNOT performs very well in NACA, Hyperelasticity and Circle, it is very expensive and its transformer-style attention fails in cases where the auxiliary input field is binary or very sparse, as it is the case with the Darcy and Maze problems where the permeability and initial pressure field are specified, respectively.
In the former, binary inputs lead to encoder embeddings that collapse into just two clusters, resulting in nearly uniform attention weights and undermining the transformer’s ability to focus selectively.
In the latter, the vast majority of zero values dominate both the attention average and its gradients, therefore, the sparse \textit{active} regions receive vanishingly small gradient updates. Such extreme cases prevent GNOT from learning meaningful patterns through the field, ultimately degrading its predictive accuracy as reflected in \Cref{tab summary} and \Cref{fig errors}.

\subsubsection{3D Case} \label{subsubsec 3d}
Handling 3D problems poses significant challenges and computational bottlenecks. The cubic scaling of memory and compute requirements, along with the overhead of FFT-based operations and the lack of mixed-precision support makes it impractical to train FNO variants at the resolutions necessary for fair comparison given our hardware constraints. Therefore, we restrict our 3D baselines to GNOT \cite{hao2023gnot} and GeomDeepONet \cite{he2024geom}, which we compare to E-UNet with two resolutions $r \in \{32, 64\}$. 

Unlike E-UNet, which does not rely on an explicit geometric representation of the domain, GeomDeepONet comprises two subnetworks: a branch that encodes geometry and load parameters, and a trunk that processes point cloud coordinates augmented by signed distance functions and sinusoidal representation (SIREN) layers. The specific hyperparameter and architecture settings used for our GeomDeepONet implementation are detailed in \Cref{list gdon}.

\begin{figure}[!t]
    \centering
    \includegraphics[width=0.99\textwidth]{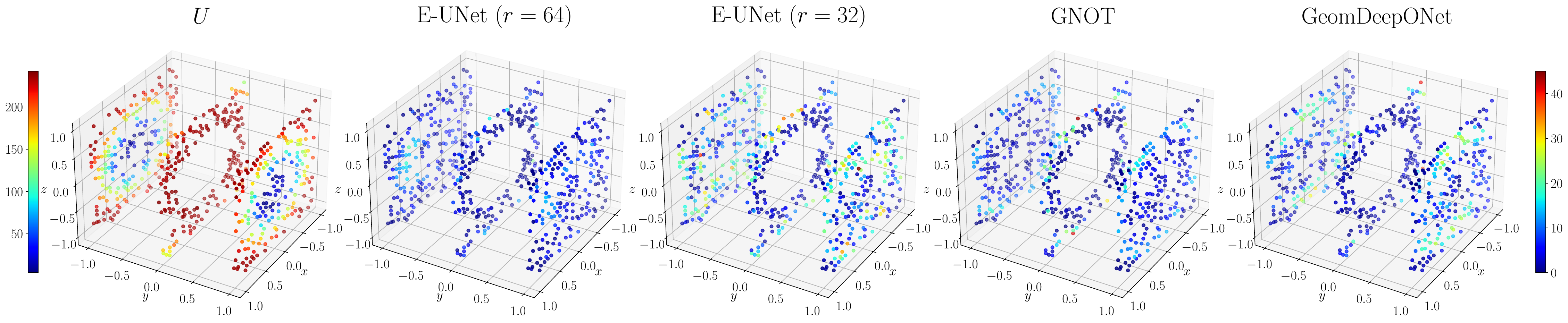} 
    \caption{\textbf{Models' absolute errors on a test sample:} The point cloud responses and errors are shown at $3$ slices $y \in [-1.05,-0.95] \cup [-0.05 , 0.05] \cup [0.95 , 1.05]$ for better visualization.} 
    \label{fig error3d}
\end{figure}

We evaluate the models on the 3D Solid example described in \Cref{subsec benchmark descriptions}. We train E-UNet, GNOT, and GeomDeepONet for $200$, $500$, and $600{,}000$ epochs, respectively. \Cref{tab 3d_summary} reports the test relative $L_2$ error ($L_2^{re}$) and training time for each model. GNOT achieves the highest predictive accuracy but requires the longest training time. 
GeomDeepONet's reliance on explicit geometric parameters does not yield a clear advantage in this setting as it underperforms even the coarse resolution E-UNet.
The reconstruction error of E-UNet reduces from $2.13E\!-\!02$  to $4.46E\!-\!03$ as the resolution increases from $32$ to $64$. 
While training E-UNet at higher resolutions could potentially further reduce $L_2^{re}$, this was not feasible under our hardware limitations. The absolute errors on a representative test sample are visualized in \Cref{fig error3d}.


\begin{table}[!hb]
    
    \centering
    \footnotesize
    \renewcommand{\arraystretch}{1.2}
    \begin{tabular}{
        |>{\centering\arraybackslash}p{2.5cm}
        |>{\centering\arraybackslash}p{2.5cm}
        |>{\centering\arraybackslash}p{2.5cm}
        |>{\centering\arraybackslash}p{2.5cm}
        |>{\centering\arraybackslash}p{2.5cm}|
    }
        \hline
        \textbf{Models} & \textbf{E-UNet $(r=64)$} & \textbf{E-UNet $(r=32)$} & \textbf{GNOT} & \textbf{GeomDeepONet} \\
        \hline
        $L_2^{re}$ & 4.48E-02 & 6.08E-02 & \textbf{3.67E-02} & 7.00E-02 \\
        Time (sec) & 2421 & 966 & 4251 & 2056 \\
        \hline
    \end{tabular}
    \caption{\textbf{Comparison of models on the 3D solid example.} The median $L_2^{\mathrm{rel}}$ and training time (in seconds) over $5$ training repetitions are reported.}
    \label{tab 3d_summary}
\end{table}

\subsection{Sensitivity Analysis} \label{subsec sensitivity analysis}
In real-world settings, data is often scarce and inevitably corrupted by noise from sensor errors and environmental disturbances, which raises practical concerns about models' reliability. To this end, we investigate how much training data is needed for the models to generalize effectively.
We also study the sensitivity of the models to perturbations in the spatial coordinates. We choose Hyperelasticity as the representative test case on which all models perform decently; its mesh nodes form unstructured point clouds, and it takes an additional input (the discretized tensile traction on top) along with each point cloud.

\begin{figure}[!hbt]
    \centering
    \vspace{-5pt}
    \includegraphics[width=0.99\textwidth]{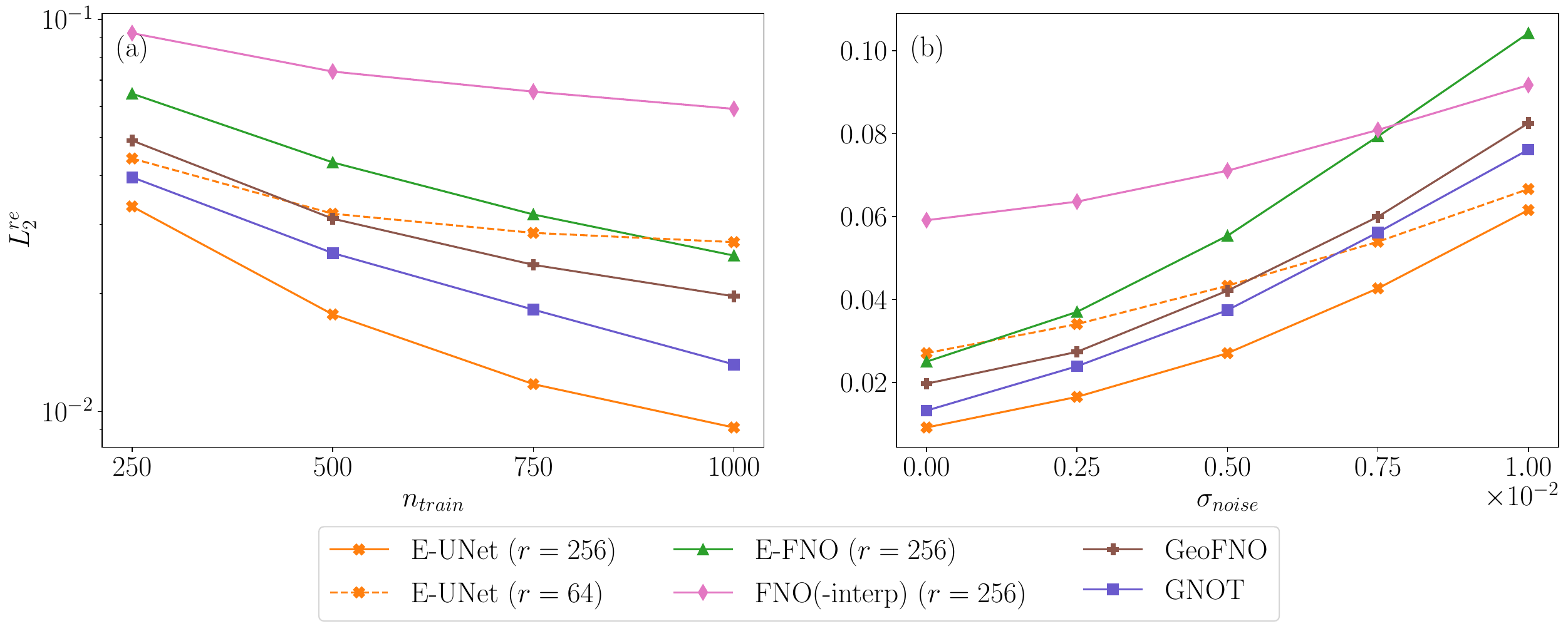} 
    \caption{\textbf{Data‐efficiency (a) and noise‐robustness (b) comparison of the models on Hyperelastcitiy example:} (a) Relative $L_2$ error on test set reduces as training samples increase for all models. High-resolution E-UNet has superior performance in limited-data conditions. (b) Additive gaussian noise on point cloud coordinates degrades the performance of the models. High-resolution E-UNet exhibits the most robustness to noise.}
    \label{fig sens}
\end{figure}

We train each models on $n_{train} \in \{ 250 , 500 , 750 ,1000\}$ samples and report the median
$L_2^{re}$ on the test set over five training repetitions.
\Cref{fig sens} demonstrates that while all models benefit from increased training data, higher-resolution E-UNet has superior performance across all training dataset sizes, indicating a stronger inductive bias and higher capacity compared to its low-resolution counterpart. In contrast, FNO variants require substantially more training data to achieve competitive performance. Notably, FNO(-interp) exhibits early performance saturation which reflects its limited capacity and inefficient topology encoding.

To assess the robustness of the models towards noise,  we perturb the training spatial coordinates confined in a unit square by additive Gaussian noise drawn from $\mathcal{N}(0 , \sigma_{noise}^2)$ where $\sigma_{noise} \in \{2.5 \! \times \! 10^{-3}  , 5\!\times\!10^{-3} , 7.5 \!\times\!10^{-3} , 10^{-2}\}$ is the prescribed standard deviation. We observe that introducing noise increases the risk of overfitting; therefore, in cases where significant overfitting occurs, we halve the number of training epochs to prevent further performance degradation. 

We present the median $L_2^{re}$'s on the noise-free test data over $5$ training repetitions in \Cref{fig sens} where we observe that high-resolution ($r=256$) E-UNet consistently outperforms other models at all noise levels. Interestingly, E-UNet's lower-resolution counterpart ($r= 64$) exhibits a smaller rate of error growth as the noise level increases. This trend can be explained as follows: looking at the slope difference at $\sigma_{noise} = 7.5\times 10^{-3}$ in \Cref{fig sens} across the two model versions, approximately $50\%$ of the points are displaced outside their original grid cells at resolution $64$, whereas at resolution $256$, nearly $95\%$ of the points fall outside their original cells during training. Such extensive displacement compromises both topology and response encodings, thus adversely impacts training.

\subsection{Inverse Mapping for Recovering Missing Points} \label{subsec inverse mapping}
Our encoding scheme provides informative representations of topology and response which can be leveraged in various mapping tasks. In real-world scenarios, sensor readings may only be available at a subset of locations while full response predictions are still required even at unknown sensor locations. To test our approach on such inverse tasks, we consider NACA example with partial observations $\boldsymbol{U}_p$ where we aim to recover the responses at all points $\boldsymbol{X}$ including the missing ones $(\boldsymbol{X}_c)$.

The core idea is to leverage the encoded partial response to estimate the full topology, which in turn enables the prediction of the full response field using our existing pretrained network. This allows us to reconstruct the full point cloud field by combining the known values with inferred estimates derived from the learned grid representations. Notably, this inverse recovery is not directly supported by many models such as GNOT, which are only designed to return predictions at the exact queried coordinates.

\begin{figure}[!t]
    \centering
    \begin{subfigure}[t]{0.73\textwidth}
        \includegraphics[width=\linewidth]{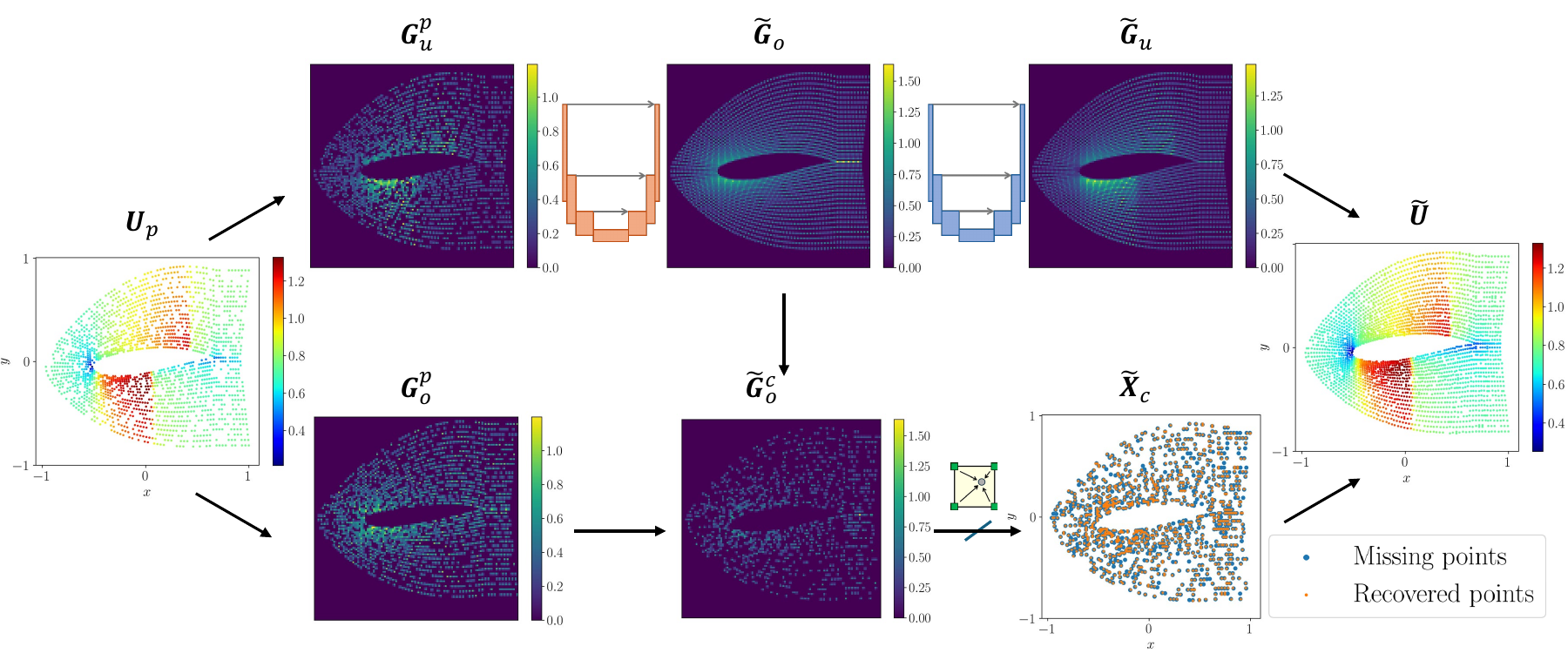}
        \caption{}
        \label{fig recovery flowchart}
    \end{subfigure}%
    \hfill
    \begin{subfigure}[t]{0.26\textwidth}
        \includegraphics[width=\linewidth]{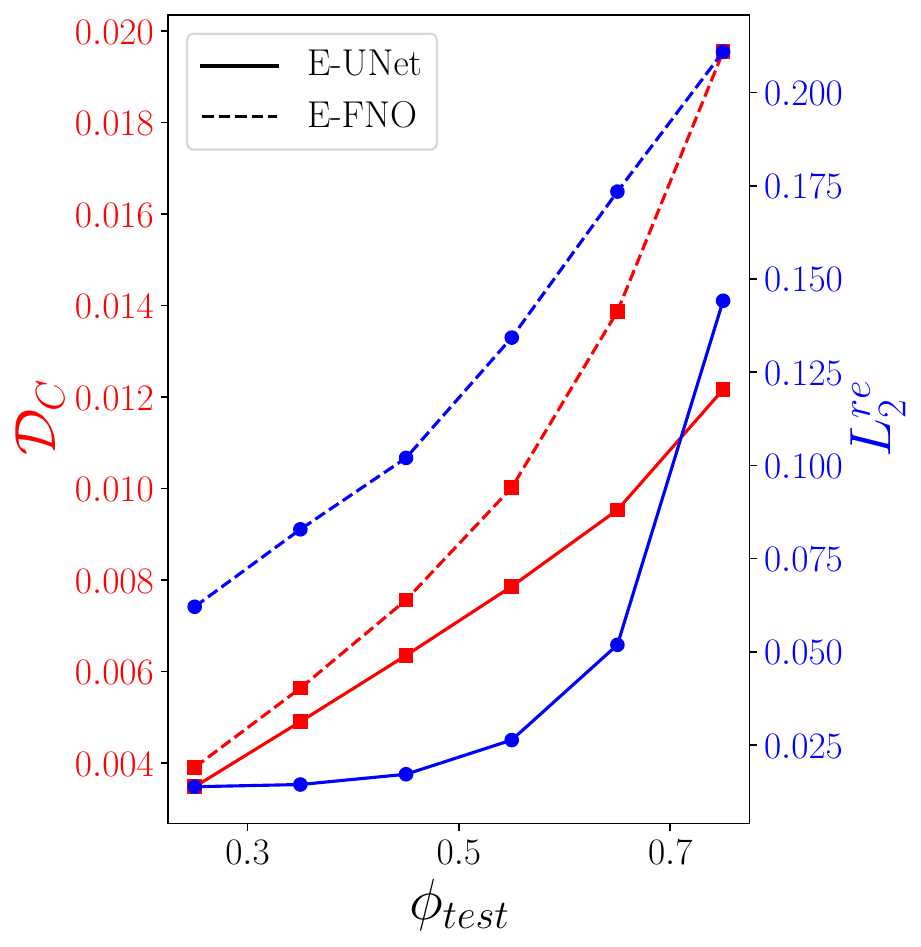}
        \caption{}
        \label{fig recovery error}
    \end{subfigure}
    \caption{\textbf{Recovery (a) framework and (b) error:} (a) A UNet is trained to map partial encoded response to the full encoded topology. The complementary encoded topology is then used to recover missing points by interpolation and thresholding. The full responses are obtained based on the approximated point cloud and the full encoded response. (b) Chamfer distance and relative $L_2$ error versus test dropout ratio $\phi_{\text{test}}$, showing low errors within the training range and a significant increase at $\phi_{test} > 0.5$ due to extrapolation.}
    \label{fig recovery}
\end{figure}

We set the dropout ratio $\phi_{train} \sim \mathcal{U}[0.2,0.5]$ to represent the fraction of missing points, i.e., $\sfrac{N-N_p}{N}$.
We encode the partial response $\boldsymbol{G}_u^p$ and the topology $\boldsymbol{G}_o^p$ with resolution $r=128$ based on the partial observations $\boldsymbol{U}_p$ the known points $\boldsymbol{X}_p$. We need to approximate the mapping between $\boldsymbol{G}_u^p$ to the encoded full topology $\boldsymbol{\widetilde{G}}_o$ via a neural network, whose output is then passed to the pretrained E-UNet from \Cref{subsubsec 2d} to perform the mapping $\boldsymbol{\widetilde{G}}_o \mapsto \boldsymbol{\widetilde{G}}_u$. 
Although various models could be used for approximating the $\boldsymbol{G}_u^p \mapsto \boldsymbol{\widetilde{G}}_o$ mapping (replacing the orange network in \Cref{fig recovery flowchart}), UNet and FNO are natural choices as they are well-suited for handling grid-structured data and have proven efficacy in the comparable tasks presented in \Cref{subsubsec 2d}.

We adopt the same architectures as in \Cref{subsubsec 2d} for both UNet and FNO without the final input-output multiplication which would otherwise zero out unintended elements. We train them by minimizing a loss similar to \Cref{eq loss3d} based on $\boldsymbol{{G}}_o$ and $\boldsymbol{\widetilde{G}}_o$ for $1000$ epochs.
Subsequently, we recover the missing points by bilinearly interpolating the encoded complementary topology $\boldsymbol{G}_o^c = \boldsymbol{\widetilde G}_o - \boldsymbol{G}_o^p$ onto $128 \times 128$ coordinate pairs evenly spaced in $[-1,1]$, from which we threshold the top $N - N_p$ points. The corresponding coordinates $\boldsymbol{\widetilde{X}}_c$ approximately match those of the missing points. The combined set $\boldsymbol{\widetilde{X}} = \boldsymbol{\widetilde{X}}_c \cup \boldsymbol{X}_p$ is then used to reconstruct the full point cloud field from $\boldsymbol{\widetilde{G}}_u$ following the same reconstruction procedure as in our original framework. This process is schematically illustrated in \Cref{fig recovery flowchart}.

We test our approach on $n_{test}=100$ samples with dropout ratios $\phi_{test} \in \{ 0.25 , 0.35 ,\cdots , 0.75\}$. We quantify the mismatch between the full approximated coordinates $\boldsymbol{\widetilde{X}}$ and the original point cloud $\boldsymbol{X}$ via Chamfer distance $\mathcal{D}_C$. That is:

\begin{equation}
\mathcal{D}_{C}(\boldsymbol{X}, \widetilde{\boldsymbol{X}}) = \frac{1}{n_{test}} \sum_{m=1}^{n_{test}}
\left[
\frac{1}{N} \sum_{n=1}^{N} \min_{\widetilde{n}} \| \boldsymbol{X}(m,n) - \widetilde{\boldsymbol{X}}(m,\widetilde{n}) \|^2
+ \frac{1}{N} \sum_{\widetilde{n}=1}^{N} \min_{n} \| \boldsymbol{X}(m,n) - \widetilde{\boldsymbol{X}}(m,\widetilde{n}) \|^2
\right]
\label{eq chamfer}
\end{equation}

In addition, we calculate the relative $L_2$ error defined in \Cref{eq rl2inv} between the target response $\boldsymbol{U}$ and its approximation interpolated on the original point cloud $\smash {\overset{ \approx}{\boldsymbol{U}}}$:

\begin{equation}
   L_2^{re} = \sqrt{\frac{\sum_{m=1}^{n_{test}}\sum_{n=1}^{N}(\boldsymbol{U}(m , n) - \boldsymbol{\overset{\approx}{\boldsymbol{U}}}(m , n))^2}{\sum_{m=1}^{n_{test}}\sum_{n=1}^{N}\boldsymbol{U}(m , n)^2}}.
    \label{eq rl2inv}
\end{equation}
The results are summarized in \Cref{fig recovery error} which shows that both metrics $\mathcal{D}_C$ and $L_2^{re}$ are relatively small for E-UNet as long as the test dropout ratio lies within the training range (i.e., $\phi_{test} < 0.5$), with values below $7\! \times\!10^{-3}$ and $2.5\!\times\!10^{-2}$, respectively. As expected, both error metrics rapidly increase beyond this range due to significant extrapolation. Remarkably, E-UNet consistently outperforms E-FNO in terms of both metrics with the performance gap widening as $\phi_{test}$ increases which indicates greater robustness of E-UNet under increasing uncertainty.


    \section {Conclusions and Future Directions} \label{sec conclusion}

In this work, we introduced a lightweight parameter-free encoder that represents point clouds as structured grids by aggregating the bilinear or trilinear footprints of each point onto its neighboring grid vertices, thereby inheriting the $O(h^\beta)$ interpolation error bounds. We then trained a customized UNet architecture to learn mappings from these enriched grid representations and benchmarked it against Fourier- and transformer-based models. We demonstrated consistent gains in accuracy, data efficiency, and noise robustness across a diverse set of 2D and 3D problems. We experimented with our powerful yet simple encoding scheme in an inverse mapping task where full point cloud responses were successfully recovered given partial observations.

While developing our encoding scheme, we encountered several challenges such as effectively handling the sparse and sharply localized nature of the encoded topology and response. These features, akin to Dirac $\delta$'s, translate to uniformly spreading energy across \textit{all} frequencies in the Fourier domain. As expected, low-mode FNOs under-resolve these sharp peaks, yet raising the frequency cutoff often destabilizes the model \cite{ovadia2023real}. In contrast, UNet which was originally developed for image segmentation, suffers much less from spectral bias and is found to be more suitable for mapping such representations.

Currently, we believe that the major limitation of our framework is its resolution dependency. Even though adopting high resolutions remains a valid reasonable option in most 2D problems, it may become infeasible and even prohibitive in 3D cases where computational cost scales cubically (instead of quadratically) with resolution. Additionally, as described in \Cref{subsec error} the reconstruction quality heavily relies on the smoothness of the underlying field. This makes it difficult for E-UNet to learn sharp gradients or discontinuities, as observed around the shock region in NACA example shown in \Cref{sec results}. We believe these shortcomings could be addressed by incorporating multi-resolution strategies that dynamically allocate representational capacity across scales. Such approaches would enable the model to resolve fine-scale features where needed while avoiding a uniform resolution increase, thus allowing computational resources to be managed more efficiently

\section*{Acknowledgments}
We appreciate the support from Office of the Naval Research (grant number $N000142312485$) and the National Science Foundation (grant numbers $2238038$ and $2211908$).

    \begin{appendices}

\setcounter{equation}{0}
\renewcommand{\theequation}{A\arabic{equation}}
\setcounter{figure}{0}
\renewcommand\thefigure{A\arabic{figure}}
\setcounter{table}{0}
\renewcommand\thetable{A\arabic{table}}
\renewcommand{\thesection}{A}

\section{Performance Statistics} \label{app stats}

We provide detailed error statistics associated with our comparative studies in \Cref{tab full stats}. These statistics are obtained for the error metric in \cref{eq rl2} across $5$ repetitions. We observe that in the majority of the cases, the mean and median values are very close and the standard deviations are relatively small for our proposed framework, E-UNet.

\begin{table}[htbp!]
    \centering
    \renewcommand{\arraystretch}{1.2}
    \small
    \begin{tabularx}{\textwidth}{|l|*{5}{>{\centering\arraybackslash}X}|}
        \hline
        \textbf{Models} & \textbf{median} & \textbf{mean} & \textbf{std} & \textbf{min} & \textbf{max} \\
        \hline
        \multicolumn{6}{|c|}{\textbf{NACA}} \\
        \hline
        E-UNet  $(r=128)$        & $1.122E\!-\!02$ & $1.131E\!-\!02$ & $4.400E\!-\!04$ & $1.077E\!-\!02$ & $1.189E\!-\!02$ \\
        E-FNO   $(r=256)$        & $1.420E\!-\!02$ & $1.417E\!-\!02$ & $5.384E\!-\!04$ & $1.360E\!-\!02$ & $1.479E\!-\!02$ \\
        FNO(-interp) $(r=256)$   & $2.400E\!-\!02$ & $2.421E\!-\!02$ & $5.201E\!-\!04$ & $2.359E\!-\!02$ & $2.479E\!-\!02$ \\
        GeoFNO                  &   $1.903E\!-\!02$     & $1.912E\!-\!02$                 &      $3.230E\!-\!03$            &      $1.609E\!-\!02$   &      $2.402E\!-\!02$            \\
        GNOT                    &      $1.207E\!-\!02$            & $1.213E\!-\!02$              &$1.586E\!-\!03$&        $1.004E\!-\!02$          &        $1.408E\!-\!02$          \\
        \hline
        \multicolumn{6}{|c|}{\textbf{Hyperelasticity}} \\
        \hline
        E-UNet   $(r=256)$       & $9.111E\!-\!03$ & $9.360E\!-\!03$ & $4.650E\!-\!04$ & $8.937E\!-\!03$ & $1.002E\!-\!02$ \\
        E-FNO  $(r=256)$      &     $2.499E\!-\!02$     &    $2.519E\!-\!02$               &     $4.690E\!-\!04$       &      $2.483E\!-\!02$         &        $2.595E\!-\!02$       \\
        FNO(-interp)       $(r=256)$      &     $5.912E\!-\!02$         &       $5.915E\!-\!02$         &     $4.690E\!-\!04$        &   $5.849E\!-\!02$            &           $5.980E\!-\!02$         \\
        GeoFNO                  &   $1.968E\!-\!02$     & $1.967E\!-\!02$                 &      $2.412E\!-\!04$            &      $1.934E\!-\!02$   &      $1.991E\!-\!02$            \\
        GNOT                    &      $1.319E\!-\!02$       & $1.335E\!-\!02$              &$6.286E\!-\!04$&        $1.284E\!-0\!2$          &        $1.438E\!-\!02$          \\
        \hline
        \multicolumn{6}{|c|}{\textbf{Darcy}} \\
        \hline
        E-UNet    $(r=128)$              &    $5.941E\!-\!03$                &        $7.737E\!-\!03$         &       $4.302E\!-\!03$          &  $5.430E\!-\!02$          &        $1.541E\!-\!02$           \\
        E-FNO    $(r=128)$            &   $9.678E\!-\!03$      &     $9.735E\!-\!03$       &       $1.656E\!-\!03$          & $7.880E\!-\!03$                 &       $1.197E\!-\!02$    \\
        FNO(-interp)   $(r=128)$    &    $1.815E\!-\!02$     & $2.019E\!-\!02$          &    $3.256E\!-\!03$    &    $1.758E\!-\!03$                 &        $2.405E\!-\!02$          \\
        GeoFNO                  &     $9.421E\!-\!03$             &  $1.333E\!-\!02$    & $1.043E\!-\!02$                 &     $7.135E\!-\!03$             &      $3.176E\!-\!02$            \\
        GNOT                    &    $2.767E\!-\!01$              &     $3.342E\!-\!01$            &  $1.281E\!-\!01$                &    $2.764E\!-\!01$              &  $5.633E\!-\!01$                \\
        \hline
        \multicolumn{6}{|c|}{\textbf{Circles $[u , v , p]^\top$}} \\
        \hline
        E-UNet  $(r = 256)$ & \makecell{$1.276E\!-\!02$\\$3.738E\!-\!02$\\$2.636E\!-\!02$} & \makecell{$1.303E\!-\!02$\\$3.885E\!-\!02$\\$2.679E\!-\!02$} & \makecell{$6.972E\!-\!04$\\$6.392E\!-\!03$\\$3.132E\!-\!03$} & \makecell{$1.255E\!-\!02$\\$3.374E\!-\!02$\\$2.222E\!-\!02$} & \makecell{$1.426E\!-\!02$\\$4.985E\!-\!02$\\$3.072E\!-\!02$} \\
        \cline{2-6}
        E-FNO   $(r = 256)$ & \makecell{$3.317E\!-\!02$\\$7.906E\!-\!02$\\$5.110E\!-\!02$} & \makecell{$3.347E\!-\!02$\\$7.930E\!-\!02$\\$5.124E\!-\!02$} & \makecell{$6.801E\!-\!04$\\$2.078E\!-\!03$\\$6.686E\!-\!04$} & \makecell{$3.286E\!-\!02$\\$7.748E\!-\!02$\\$5.048E\!-\!02$} & \makecell{$3.443E\!-\!02$\\$8.262E\!-\!02$\\$8.262E\!-\!02$} \\
        \cline{2-6}
        FNO(-interp) $(r = 256)$& \makecell{$5.535E\!-\!02$\\$1.201E\!-\!01$\\$4.799E\!-\!02$} & \makecell{$5.539E\!-\!02$\\$1.206E\!-\!01$\\$4.780E\!-\!02$} & \makecell{$1.274E\!-\!03$\\$2.009E\!-\!03$\\$6.216E\!-\!04$} & \makecell{$5.363E\!-\!02$\\$1.186E\!-\!01$\\$4.687E\!-\!02$} & \makecell{$5.712E\!-\!02$\\$1.231E\!-\!01$\\$4.853E\!-\!02$} \\
        \cline{2-6}
        GeoFNO   & \makecell{$1.028E\!-\!01$\\$1.965E\!-\!01$\\$8.531E\!-\!02$} & \makecell{$2.054E\!-\!01$\\$3.809E\!-\!01$\\$1.749E\!-\!01$} &\makecell{$1.854E\!-\!01$\\$3.293E\!-\!01$\\$1.506E\!-\!01$} & \makecell{$4.417E\!-\!02$\\$8.721E\!-\!02$\\$5.584E\!-\!02$} & \makecell{$4.776E\!-\!01$\\$8.482E\!-\!01$\\$4.028E\!-\!01$} \\
        \cline{2-6}
        GNOT     & \makecell{$1.508E\!-\!02$\\$3.134E\!-\!02$\\$1.558E\!-\!02$} & \makecell{$1.554E\!-\!02$\\$3.193E\!-\!02$\\$1.593E\!-\!02$} & \makecell{$1.073E\!-\!03$\\$2.023\!-\!03$\\$9.922E\!-\!03$} & \makecell{$1.496E\!-\!02$\\$2.999E\!-\!02$\\$1.505E\!-\!02$} & \makecell{$1.745E\!-\!02$\\$3.519E\!-\!02$\\$1.749E\!-\!02$} \\
        \hline
        \multicolumn{6}{|c|}{\textbf{Maze}} \\
        \hline
        E-UNet      $(r=256)$       &        $2.724E\!-\!01$          &        $2.797E\!-\!01$        &         $1.673E\!-\!02$         &  $2.715E\!-\!01$   &        $3.096E\!-\!01$          \\
        E-FNO        $(r=128)$           &        $7.146E\!-\!01$          &    $7.129E\!-\!01$        &        $7.493E\!-\!03$           &         $7.012E\!-\!01$         &       $7.213E\!-\!01$             \\
        FNO(-interp)     $(r=256)$        &     $8.145E\!-\!01$             &        $8.156E\!-\!01$         &         $7.186E\!-\!03$         &         $8.083E\!-\!01$         &      $8.257E\!-\!01$            \\
        GeoFNO                  &    $7.604E\!-\!01$        &        $7.625E\!-\!01$        &    $6.467E\!-\!03$              &      $7.550E\!-\!01$            &   $7.707E\!-\!01$               \\
        GNOT                    &     $9.941E\!-\!01$             &     $9.941E\!-\!01$            &  $6.148E\!-\!07$                & $9.941E\!-\!01$                 &   $9.941E\!-\!01$               \\
        \hline

        \multicolumn{6}{|c|}{\textbf{Solid 3D}} \\
        \hline
        E-UNet $(r=64)$                 &       $4.479E\!-\!02$       &   $4.575E\!-\!02$              &         $2.381E\!-\!03$         &   $4.364E\!-\!02$               &       $4.972E\!-\!02$           \\
        GNOT                  &       $3.668E\!-\!02$            &  $3.692E\!-\!02$               &    $1.001E\!-\!03$              &    $3.586E\!-\!02$              &  $3.847E\!-\!02$                \\
        GeomDeepONet            &   $6.998E\!-\!02$               &    $7.019E\!-\!02$             &     $3.170E\!-!03$             &           $6.657E\!-\!02$       &          $7.477E\!-\!02$        \\
        \hline
    \end{tabularx}
    \caption{Performance statistics of models across benchmark problems. Each cell in the Circles block shows values for $u$, $v$, and $p$. Performance statistics of models across benchmark problems. Each cell in the Circles block shows values for $u$, $v$, and $p$.}
    \label{tab full stats}
\end{table}

\pagebreak

\setcounter{equation}{0}
\renewcommand{\theequation}{A\arabic{equation}}
\setcounter{figure}{0}
\renewcommand\thefigure{B\arabic{figure}}
\setcounter{table}{0}
\renewcommand\thetable{B\arabic{table}}
\renewcommand{\thesection}{B}

\section{Models' Configurations} \label{app model config}

We provide additional details on models' configuration and hyperparameters used for problems presented in \Cref{sec results}. \Cref{tab problem_parameters} presents E-UNet's parameters used to obtain the results in \cref{tab summary}. \Cref{list gnot,list gdon} list the arguments and settings used in the original code implementations of GNOT and GeomDeepONet, respectively.

\begin{table}[htbp!]

    \centering
    
    \begin{tabular}{|l|c|c|c|c|c|c|}
    \hline
    Problem & $c_{\text{in}}$ & $c$ & $c_{\text{out}}$ & $r$ & $L$ & \# Params \\
    \hline
    NACA & 1 & 32 & 1 & 128 & 4 & 13.34M \\
    Hyperelasticity & 2 & 16 & 1 & 256 & 5 & 13.37M \\
    Darcy & 2 & 16 & 1 & 128 & 5 & 13.37M \\
    Circles & 2 & 16 & 3 & 256 & 5 & 13.57M \\
    Maze & 2 & 16 & 1 & 256 & 5 & 13.37M \\
    3D Solid & 1 & 12 & 1 & 64 & 4 & 3.18M \\
    \hline
    \end{tabular}
    \caption{UNet configuration for each problem}
    \label{tab problem_parameters}
\end{table}

\begin{lstlisting}[language=Python, caption={GNOT Configuration},label={list gnot}]
# GNOT Configuration
--dataset         'elas2d'      # Options: naca2d, darcy2d, solid3d, ...
--component       'all'
--seed            12
--gpu             0
--use-tb          1
--comment         ""
--train-num       1000
--test-num        100
--sort-data       0
--normalize_x     'none'
--use-normalizer  'none'

# Training
--epochs          500
--optimizer       'AdamW'
--lr              0.001
--weight-decay    5e-6
--grad-clip       1000.0
--batch-size      4
--val-batch-size  100
--no-cuda         False

# Learning rate schedule
--lr-method       'cycle'
--lr-step-size    50
--warmup-epochs   50

# Loss and Model
--loss-name       'rel2'
--model-name      'GNOT'           # Options: CGPT, GNOT
--n-hidden        64
--n-layers        3

# MLP and Attention
--act             'gelu'
--n-head          4
--ffn-dropout     0.0
--attn-dropout    0.0
--mlp-layers      3
--attn-type       'linear'
--hfourier-dim    0

# GNOT-specific
--n-experts       1
--branch-sizes    2
--n-inner         4

\end{lstlisting}

\begin{lstlisting}[language=Python, caption={GeomDeepONet Configuration},label={list gdon}]
# GeomDeepONet Configuration
--model           'DeepONetCartesianProd'
--data-type       'float32'
--n-load-paras    9
--n-comp          1
--hidden-dim      32
--scale           2             # Scales HIDDEN dimension
--n-out-pt        5000
--batch-size      16
--epochs          600000
--learning-rate   2e-3

# Architecture
--activation      'swish'       # For standard layers
--siren-w0        10.0          # For sinusoidal layers
--outNet-layers   [50, 50, 32]  # Scaled by 'scale' parameter
--outNet2-layers  [64, 64, 32]  # Uses sinusoidal representation
--geoNet-layers   [50, 50, 32]  # Scaled by 'scale' parameter 
--geoNet2-layers  [64, 64, 32]  # Standard activation

# Data Processing
--data-split      'distance'    # Distance-based train/test split
--data-limit      2500          # Number of samples used
--output-clip     [0, 310]      # Clip stress values
--use-scaler      'MinMaxScaler'

# Optimizer
--optimizer       'adam'
--decay           'inverse_time'
--decay-factor    0.1

\end{lstlisting}

\end{appendices}
    \pagebreak
    \bibliographystyle{unsrt} 
    \bibliography{01_Ref.bib}     
\end{document}